\documentclass[acmlarge]{acmart}
 \usepackage{orcidlink}
\usepackage[marginal]{footmisc}

\AtBeginDocument{%
  \providecommand\BibTeX{{%
    \normalfont B\kern-0.5em{\scshape i\kern-0.25em b}\kern-0.8em\TeX}}}

\setcopyright{none}

\acmDOI{10.1145/3699747}

\usepackage{xspace}
\usepackage{todonotes}
\usepackage{xcolor}
\usepackage{multirow}
\usepackage{array}

\usepackage{tcolorbox}
\tcbset{boxrule=0pt, arc=8pt, outer arc=8pt, colback=pink!90, colframe=pink!90, boxsep=2pt, left=2pt, right=2pt, top=2pt, bottom=1pt, on line}

\begin{document}

\title{Sensor2Text: Enabling Natural Language Interactions for Daily Activity Tracking Using Wearable Sensors}

\author{Wenqiang Chen*\,\orcidlink{0000-0001-8167-6843}}
\affiliation{%
  \institution{Chinese Academy of Sciences, SIAT}
  \institution{Massachusetts Institute of Technology}
  \country{USA}
}
\email{wenqiang@mit.edu}

\author{Jiaxuan Cheng*\,\orcidlink{0009-0003-5558-9887}}
\affiliation{%
  \institution{Massachusetts Institute of Technology}
  \country{USA}
}

\author{Leyao Wang\,\orcidlink{0009-0000-0958-5605}}
\affiliation{%
  \institution{Vanderbilt University}
  \country{USA}
}

\author{Wei Zhao\,\orcidlink{0000-0002-6268-2559}}
\affiliation{%
  \institution{University of SIAT, Shenzhen}
  \country{China}
}

\author{Wojciech Matusik\,\orcidlink{0000-0003-0212-5643}}
\affiliation{%
  \institution{Massachusetts Institute of Technology}
  \country{USA}
}

\renewcommand{\shortauthors}{Chen, et al.}

\begin{abstract}

Visual Question-Answering, a technology that generates textual responses from an image and natural language question, has progressed significantly. Notably, it can aid in tracking and inquiring about daily activities, crucial in healthcare monitoring, especially for elderly patients or those with memory disabilities. However, video poses privacy concerns and has a limited field of view. This paper presents Sensor2Text, a model proficient in tracking daily activities and engaging in conversations using wearable sensors. The approach outlined here tackles several challenges, including low information density in wearable sensor data, 
insufficiency of single wearable sensors in human activities recognition,  
and model's limited capacity for Question-Answering and interactive conversations.
To resolve these obstacles, transfer learning and student-teacher networks are utilized to leverage knowledge from visual-language models. Additionally, an encoder-decoder neural network model is devised to jointly process language and sensor data for conversational purposes. Furthermore, Large Language Models are also utilized to enable interactive capabilities. The model showcases the ability to identify human activities and engage in Q\&A dialogues using various wearable sensor modalities. It performs comparably to or better than existing visual-language models in both captioning and conversational tasks. To our knowledge, this represents the first model capable of conversing about wearable sensor data, offering an innovative approach to daily activity tracking that addresses privacy and field-of-view limitations associated with current vision-based solutions.
\end{abstract}



\begin{teaserfigure}
\centering
    \includegraphics[width=0.95\textwidth]{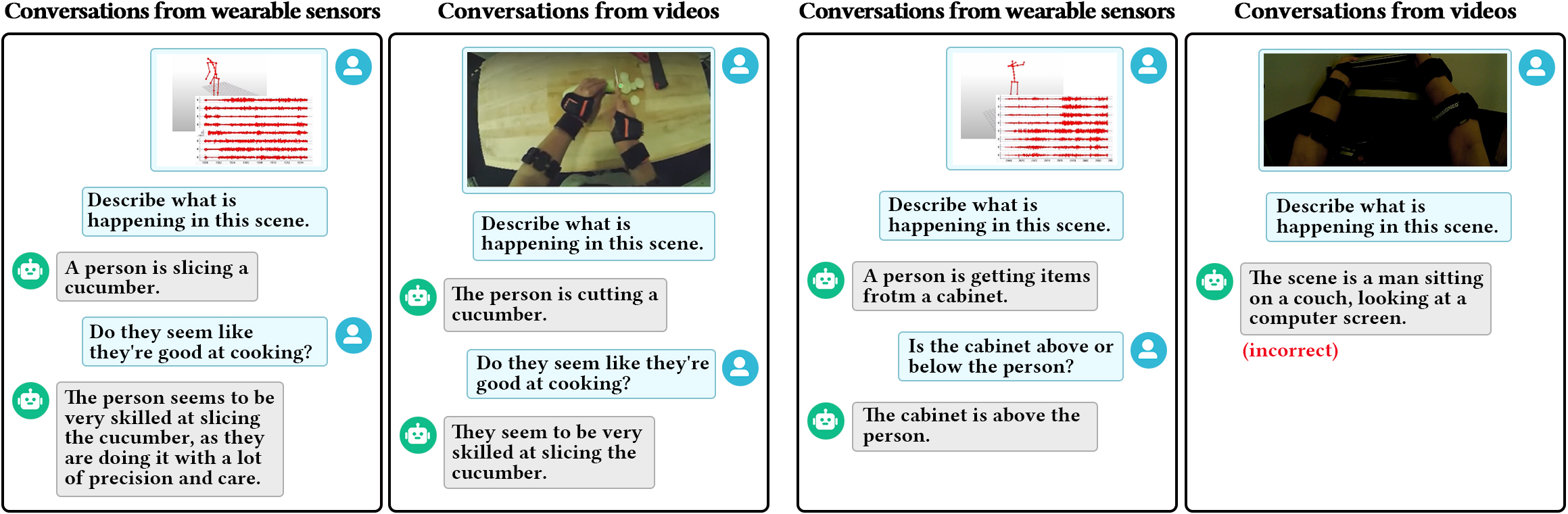}
    \caption{Sensor2Text demonstrates an understanding of wearable sensor inputs and interactive capability, compared with conversations generated from existing vision models on the ground truth videos. Sensor2Text accurately comprehends daily activities through wearable sensor data,  circumventing video-related constraints like inadequate lighting or obstructed viewpoints.}
    \Description{}
    \label{fig:teaser}
\end{teaserfigure}

\footnote{* The first and second authors contributed equally to this work.}



\maketitle
\section{Introduction}

Visual Question Answering is an exciting task in computer vision and natural language processing that has gained much progress in recent years; it involves generating a text response given an image and natural language question \cite{VQA}. Notably, visual question-answering can be applied in tracking personal daily activities, which shows great promise for healthcare monitoring, especially for elderly people or patients with Alzheimer's. While traditional monitoring systems can only send raw footage generated by monitoring cameras to caregivers, current question-answering systems enable conversational interactions with the users. This capability allows caregivers to inquire about various aspects of the target's well-being, such as medication adherence, dietary habits, sleep patterns, and more.

However, using video as the input modality to question-answering has several issues. Video suffers from obstructions, restricted viewpoints, and vulnerability to inadequate lighting, all of which are common occurrences in daily activities monitoring. More importantly, using constant video surveillance has privacy concerns and may make the subject feel uncomfortable. Therefore, it is crucial to devise an innovative approach to replace the reliance on cameras, which enables the collection of data on subjects' daily activities while also preserving their privacy.   

It is observed that sensor-based data, particularly those obtained from wearable sensors, can be a promising alternative. Wearable sensors offer several advantages over video surveillance. Firstly, they allow the subject to roam freely outside of designated areas, as opposed to video which requires the subjects to remain within the camera coverage zones. Additionally, compared with video cameras, wearable sensors can provide more information regarding one's health and physical activity, making it a better option for healthcare monitoring. Finally, they are less invasive in privacy compared to video cameras.

In this paper, we propose Sensor2Text, a sensor language model capable of conditioning itself on wearable sensor data to perform the aforementioned question-answering, illustrated in Figure \ref{fig:teaser}. Sensor2Text is interactive and versatile for subjects and caregivers to track daily activities. Additionally, it functions without requiring any visual input, making it less obtrusive than video monitoring and applicable in dark or occluded settings. Furthermore, compared with existing video and sensor captioning models \cite{xu2023mplug2, luo2020univl, fan2020inhome}, Sensor2Text stands out for its chat capabilities and proficiency in question-answering tasks, enhancing its utility in tracking daily activities.

However, developing a sensor-language model like Sensor2Text presents several challenges. Firstly, wearable sensors lack the density of information found in video data regarding human activities. To overcome this, we employ cross-modality supervision using teacher-student networks. In this approach, a model trained on visual camera data acts as a teacher, efficiently transferring its knowledge to models analyzing sensor data after relatively few sensor examples.

In addition, a single wearable sensor is often insufficient when differentiating between similar activities. To fix this, we use multiple wearable devices to track human activities. We jointly process the data modalities using a multi-modal encoder-decoder neural network architecture and align the temporal dynamics between the modalities with time-dependent output tokens.

Furthermore, even if the model can comprehend sensor data, it lacks the capacity for Question-Answering or interactive conversations. To tackle this obstacle, Large Language Models are integrated into Sensor2Text to endow it with conversational abilities.

We conducted a comprehensive evaluation of the Sensor2Text model, assessing its performance through both qualitative and quantitative analyses. The results demonstrate that Sensor2Text exhibits robust capabilities in responding accurately to queries and providing detailed descriptions based on wearable sensor data input. In our tests, Sensor2Text achieved captioning scores comparable to those of leading video-language models, with an average BLEU score of 73.0 and a CIDEr score of 62.2. This closely aligns with the scores of existing vision-language models, which average a BLEU score of 76.7 and a CIDEr score of 62.5, underscoring its competitive sensor recognition capabilities. Moreover, Sensor2Text shows remarkable performance in challenging conditions, such as low-light or occluded environments where vision-based models are ineffective. Furthermore, it demonstrates strong generalization across different users and effective utilization of multiple sensor data modalities. Additionally, ablation studies were conducted to ascertain the contributions of individual components to the model’s overall efficacy and training methodology.

In summary, the major contributions of our work are:
\begin{itemize}

\item To the best of our knowledge, we are the first to create a chatbot capable of engaging in Q\&A and conversations regarding various human daily activities, achieved through comprehension of multi-modal wearable sensor data. 

\item We design a transformer-based neural network encoder, including late fusion and a 1D convolutional tokenizer, and use auxiliary training losses to extract useful sensor representations while efficiently transferring knowledge from video.

\item We evaluate our model through extensive experiments, including qualitative and quantitative assessments, generalization tests on unseen subjects, and ablation studies.

\end{itemize}
\section{Related Work}

Our study is closely connected to research within the domains of Human Activity Recognition, Video Captioning, and Multi-modal Large Language Models. Below, we provide an in-depth exploration of these areas, elucidating how Sensor2Text leverages insights from prior research to pioneer a novel approach enabling conversations and question-answering about content in wearable sensors. This innovative approach mitigates privacy concerns and viewpoint limitations in existing vision-based solutions, making it a promising alternative in applications such as healthcare monitoring.

\subsection{Human Activity Recognition}

The domain of Human Activity Recognition (HAR)\cite{chen2024cavatar}, which focuses on identifying human actions through sensor data analysis, has witnessed notable progress in recent years, particularly in leveraging machine learning techniques for interpreting human activities from sensor data. Researchers have utlized machine learning techniques such as kernel discriminant analysis and Long Short-Term Memory (LSTM)-based Neural Structured Learning \cite{uddin2021human, 10.1145/3090076}, Decision Trees, Support Vector Machines, and Neural Networks \cite{misc_human_activity_recognition_using_smartphones_240} to recognize activities from smartphone data. Another area of interest involves the use of radio signals interacting with human bodies to comprehend movements and behaviors. Specifically, previous studies have demonstrated the use of radio signals for tasks such as location tracking \cite{WiTrack, fan2020inhome, 10.1145/3130940}, extraction of 3D skeletons of nearby individuals \cite{10.1145/3230543.3230579}, and action classification \cite{li2019making}. Finally, some studies also employ wearable sensors and deep learning techniques for HAR and other related tasks \cite{10.1145/3517252, 10.1145/3550331, 10.1145/3550316, 10.1145/3130902, chen2021sensecollect, chen2017floc, huang2019g, chen2017virtual, chen2020smartwatch, wu2020power, chen2020continuous, guan2019faceinput, chen2022making, chen2022viwatch, chen2021viobject, chen2019low, chen2019taprint, chen2018vitype, chen2021vifin, chen2023robust, }, including unsupervised domain adaptation \cite{10.1145/3380985}, contrastive predictive coding \cite{10.1145/3463506}, and deep unsupervised clustering \cite{10.1145/3448074}.

While recent studies have made significant progress in Human Activity Recognition (HAR), the predominant focus lies in classifying recognized activities. However, Sensor2Text is capable of performing captioning and utilizing information from multiple modalities of wearable sensor data. Additionally, our approach endeavors to perform question-answering and have conversations regarding the content gleaned from wearable sensor data during daily activities rather than only simple text messages from captioning.

\subsection{Video Captioning}
Video captioning refers to the generation of descriptive text for video content. One strategy for video captioning involves applying image captioning models to uniformly sampled frames \cite{ghaderi2022diverse, 10272675}. Recent advancements employ recurrent neural networks (RNNs) \cite{venugopalan2015sequence}, attention mechanisms \cite{vaswani2023attention}, hierarchical architectures \cite{ye2022hierarchical, wang2016hierarchical} to capture more temporal intricacies in the video. Furthermore, recent work studies the integration of audio and video comprehension through advancements in multi-modal learning \cite{chen2023vast, xu2023mplug2}.

Another development pertinent to video captioning is Vision-Language Pre-training (VLP), which aims to enhance the performance of multi-modal foundation models across various vision-language tasks through unsupervised learning. For instance, CLIP \cite{radford2021learning} utilizes conservative loss between image/text pairs to pre-train a model on internet-scraped data. These models often adopt architectures like dual-encoders \cite{ni2021large, radford2021learning, xu2021e2evlp}, fusion-encoders \cite{tan2019lxmert}, or a hybrid of both as encoder-decoder architectures \cite{li2023blip2}. VLP models can be applied across a range of vision-based tasks, including classification and captioning, among others.

While great advancements have been made in video captioning, our work, in contrast, is intended to employ wearable sensor data for conversational interactions with the user. However, the sensor-language model continues to leverage advancements in video captioning and VLP by employing visual language models in transfer learning to distill the learned knowledge to wearable sensor data. After training, the user will be able to conduct Q\&A conversations using wearable sensor data as input, without requiring any visual input.

\subsection{Multi-modal Large Language Models}
Large Language Models (LLMs) have garnered significant attention in recent years for their ability to comprehend and generate human-like text. There is a growing interest in extending their capabilities to process modalities of inputs other than language. One approach is to use LLMs as controllers, with existing multi-modal models serving as tools \cite{brown2020language, 10.1145/3643540, 10.1145/3643505, niu2024screenagent, kim2024healthllmlargelanguagemodels}. In this setup, the LLM decides which tools to call based on the user's text instruction and incorporates results from these multi-modal models. Other work focuses on developing fundamental multi-modal LLMs by aligning the representation space of pre-trained unimodal models with that of LLMs, allowing the language model to natively understand other data modalities rather than through a proxy. Models like BLIP2 \cite{li2023blip2} and LLaVA \cite{liu2023visual} exemplify this approach.

Despite these advancements, the majority of multi-modal LLMs primarily accept images as input \cite{yin2023survey}. Some studies have explored processing other data modalities, such as 3D point clouds \cite{guo2023pointbind} or audio \cite{zhang2023videollama}, but few have attempted to extend the capabilities of LLMs to comprehend wearable sensor data. Our proposed model, Sensor2Text, aims to bridge this gap by enabling the understanding of wearable sensor data and its integrate them withLLMs.

In summary, our research extends the existing body of literature on Human Activity Recognition, Video Captioning, Vision-Language Pre-training (VLP), and multi-modal Large Language Models. What sets our model apart is its utilization of wearable sensor data, rather than conventional visual data, to perform Q\&A and chat about an individual's daily activities.
\section{Preliminary Study}

This section first explores the capability of wearable sensor data in recognizing and distinguishing daily activities. It then assesses the feasibility of training such a sensor model due to the scarcity of comprehensive multi-modal wearable sensor datasets.

\subsection{Sensor Characteristics for Activity Tracking}

\begin{figure}
\centering
\includegraphics[width=1\textwidth]{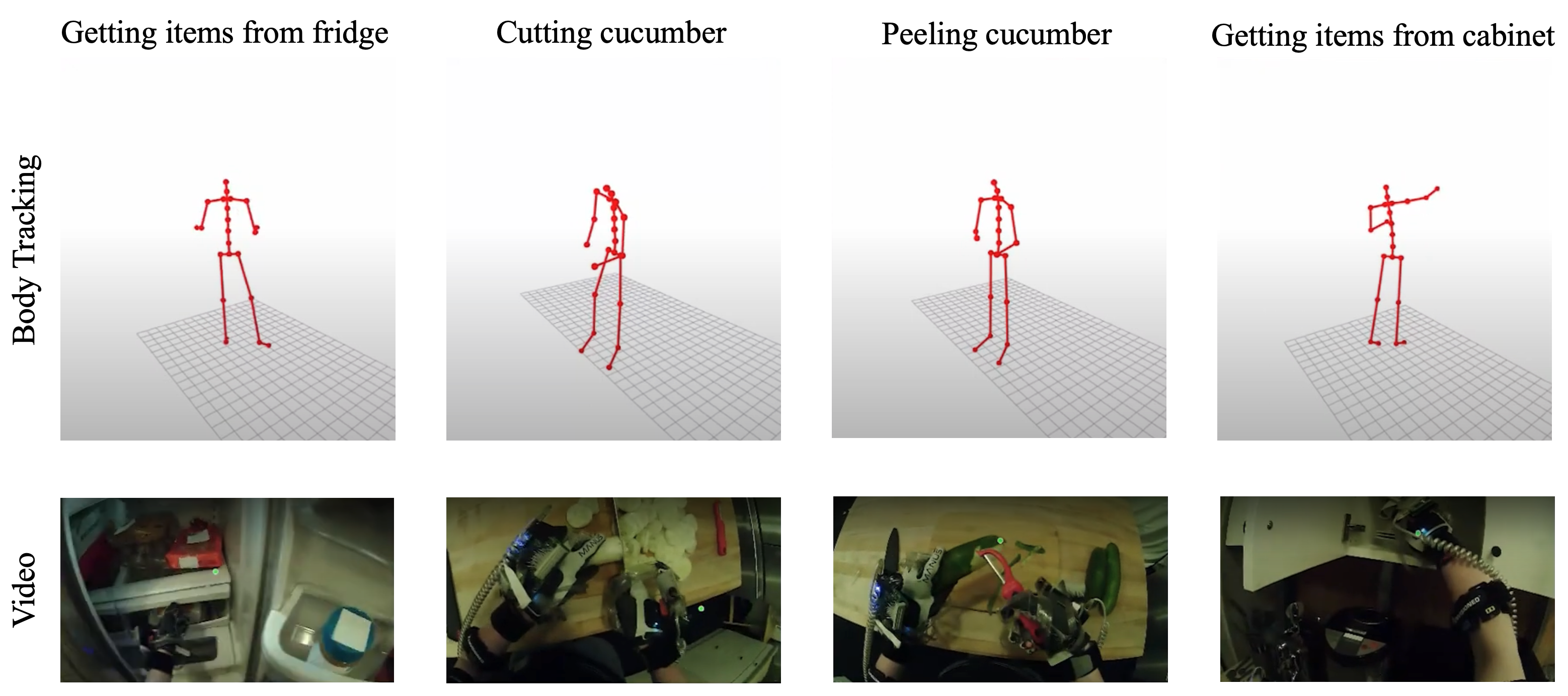}
\caption{Visualization of body tracking data using wearable sensors. Different activities display distinct body positions, demonstrating the feasibility of using wearable sensors to identify daily activities.}
\label{fig:characteristics}
\end{figure}

We observe that wearable sensor data can effectively be used to identify human activities for the following reasons. Firstly, activities that may appear similar through video cameras often yield distinct sensor readings. Illustrated in Figure \ref{fig:characteristics}, while actions like peeling or cutting cucumbers may seem alike in video images, the corresponding human skeleton representations derived from sensor data exhibit notable differences.

Additionally, wearable sensors reveal dynamic variations. Each wearable sensor gathers time series data rather than static snapshots, capturing a subject's movement patterns over time that can further aid in distinguishing similar activities.

Furthermore, integrating multiple modalities from wearable sensors enhances the differentiation of human activities. While certain activities may exhibit similar body tracking readings, they may yield entirely different readings from other wearable sensors. For instance, although body tracking data excels in distinguishing activities with various movements, it may struggle to discern certain nuances, such as differentiating between cutting a cucumber and a potato. However, by integrating muscle activation sensors that monitor electromyography signals, it becomes possible to differentiate between the objects the subject is holding \cite{10.1145/3643547, FLIGGE2013402}, effectively distinguishing the action of cutting a potato from cutting a cucumber. By processing multiple modalities of sensor data through a multi-modal deep learning architecture, it becomes possible to make use of each modality in determining the details of the activity.

To summarize, through the integration of diverse and carefully selected data modalities alongside appropriate temporal-aware processing, it is possible that sensor data can be utilized to classify daily activities with performance equal to, or even surpassing, that achieved with video data.

\subsection{Feasibility}

A significant concern regarding the feasibility of constructing a sensor-language model is the scarcity of large-scale wearable sensor datasets. In contrast, visual-language models benefit from abundant datasets, facilitating extensive training or fine-tuning for robust performance in captioning tasks. Consequently, an insightful approach emerges: we can transfer the knowledge acquired from training visual-language models to sensor data through transfer learning and teacher-student networks. 

Indeed, prior research \cite{guo2023pointbind, han2023imagebindllm}  demonstrates the feasibility of leveraging large visual-language datasets for downstream fine-tuning by aligning the encoder of new modality data with an existing visual encoder. This significantly mitigates the need for an extensive amount of wearable sensor data. Accordingly, employing wearable sensor data to train a sensor language model is feasible.

\section{Method}

\begin{figure}
    \centering
    \includegraphics[width=0.98\linewidth]{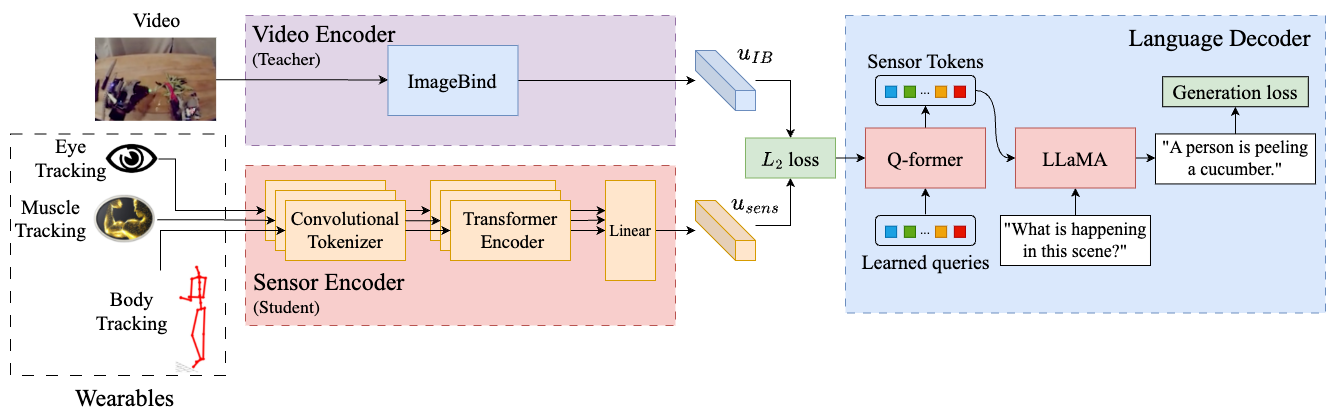}
    \caption{Architecture diagram of Sensor2Text. }
    \label{fig:architecture}
\end{figure}

Sensor2Text is a novel model designed to generate textual responses based on sensor data and textual prompts. The primary challenges in developing such a model are the low information density of single modal sensor data and the limited sizes of existing sensor datasets. To address these challenges, a unique architecture is proposed, illustrated in Figure \ref{fig:architecture}. The model consists of two main subcomponents: a sensor encoder and a language decoder. The sensor encoder takes multiple modalities of sensor data as input and outputs a learned representation vector that incorporates information from all sensors. The language decoder then takes this representation along with the language prompt and generates the textual response.

To efficiently train the model despite the lack of sensor data, a vision encoder is introduced as a teacher network to train the sensor encoder. Additionally, transfer learning techniques are employed to train the language decoder. To facilitate the training process, a dataset is carefully selected to include videos, labels, and multiple sensor data modalities.

The following subsections provide a detailed description of the architecture, including the sensor encoder and language decoder, as well as the procedure used to train both components.

\subsection{Sensor Encoder}
This section introduces the sensor encoder, which extracts useful features from time series data collected by the various wearable sensors and integrates them into a single representation.

We utilize distinct encoders to extract features from each sensor data modality separately before combining them with late fusion. A transformer encoder \cite{vaswani2023attention} is applied to each data modality due to its excellent performance in capturing intricate dependencies within the sequential data. Each sequence of wearable sensor data, denoted as $x^{(i)} \in \mathbb{R}^{T_i \times d_i}$ for the $i$th sensor modality, where  $d_i$ denotes the number of features and  $T_i$ represents the number of timesteps, is processed via a 1D convolutional tokenizer. This converts the raw sensor data into tokens that are more easily ingested by the downstream transformer. The sequential data is segmented into fixed-length windows and a learned 1D convolution is applied to each window. The resulting token for each window is represented as $u^{(i)}_j = \theta_{\text{tok}}^\top x^{(i)}_{s*j,\ldots,s*j+d-1}$, where $d$ indicates the window size, $s$ denotes the stride between consecutive windows, $\theta_{\text{tok}} \in \mathbb{R}^{d \times d_\text{encoder}}$ represents the learnable weights of the tokenizer, and $u^{(i)}_j \in \mathbb{R}^{d_\text{encoder}}$ represents the output tokens.

The tokenizer then applies a sine/cosine positional encoding onto each token to inject temporal information into each token, rendering different positions distinguishable to the downstream transformer. The positional encoding for the $j^{th}$ position and $d^{th}$ dimension is given as $\text{PE}(j,d)$, where:

\[
\begin{aligned}
\text{PE}(j, 2k) &= \sin(j / 10000^{2k/d_\text{encoder}}) \\
\text{PE}(j, 2k+1) &= \cos(j / 10000^{2k/d_\text{encoder}})
\end{aligned}
\]

Each positional encoding is directly added onto the token sequence for each input data modality, resulting in a modified sequence of position-aware tokens denoted as \( u'_j = u_j + \text{PE}(j) \). A learnable classification token, denoted as \texttt{[CLS]}, is prepended to this sequence. After processing in the downstream sequence-to-sequence transformer, the output of this token will serve as the single representation for the entire sequence. Consequently, we obtain a sequence of tokens \( \textbf{u} = (\text{[CLS]}, u'_1, ..., u'_n) \).

Next, the position-aware tokens are processed in a transformer encoder to obtain an intermediate representation for each sensor. The transformer encoder consists of $N$ layers of alternating self-attention and feedforward layers, effectively capturing both long and short-term dependencies in the sensor tokens. The output vector of the \texttt{[CLS]} token is extracted to obtain an intermediate representation for each sensor modality, $y^{(i)} = \text{Transformer}(\textbf{u}^{(i)})$, where $y^{(i)} \in \mathbb{R}^{d_\text{encoder}}$.

Finally, the late fusion technique is employed to obtain the ultimate representation vector. This process involves concatenating the representation vectors of each sensor and then applying a final feedforward layer to capture potential interdependencies among all sensor modalities. The outcome is a single representation vector, \( y_{\text{sens}} \in \mathbb{R}^{d_{\text{output}}} \), which encapsulates all relevant information from the wearable sensor data, where \( d_{\text{output}} \) is the dimension of the final representation vector.

\subsection{Auxiliary Training Objective}
We leverage a visual encoder as a teacher network to train the sensor encoder. Through this method, the sensor encoder can effectively distill valuable information from the visual encoder while avoiding the instability, high computational cost, and susceptibility to overfitting that comes from training the encoder with an end-to-end text generation loss. For this purpose, the ImageBind model \cite{girdhar2023imagebind} is utilized as the designated teacher network. ImageBind has the capacity to encode diverse modalities such as visual, audio, IMU, thermal, depth, and text data into a unified embedding space. Its proficiency in learning representations that transcend modality-specific boundaries makes it an ideal candidate to act as a teacher network.

Specifically, the number of output features of the sensor encoder, $d_\text{encoder}$, is set equal to the output dimension of ImageBind and train the sensor encoder to align with the visual encoder using $L_2$ loss on paired sensor/video examples. Adhering to the methodology adopted in ImageBind for video processing, two representative frames are selected for each video for processing. For video $i$, denoted as $V_i = [v_0; ...; v_{n_i-1}]$, where $n_i$ is the number of frames, we select $v_0$ and $v_{n_i/2}$ to be inputted into ImageBind. The total loss is formulated as follows:

\[
\mathcal{L}_{\text{aux}}(x; \theta) = \sum_{(x_i, V_i) \in \mathcal{D}} \left\| \text{Encoder}(x_i; \theta) - \text{ImageBind}(V_{i,0}, V_{i, n_i/2}) \right\|^2 + R(\theta)
\]
Here, $x_i$ and $V_i$ represent paired sensor and video data from the dataset $\mathcal{D}$, with $\theta$ denoting the encoder's weights and R denoting a regularization term. Through this approach, the sensor encoder can generate meaningful representations without necessitating the execution of the entire Sensor2Text model in an end-to-end manner. After this training phase, the weights of the sensor encoder are frozen for the duration of the remaining training process.

\subsection{Language Decoder}

After acquiring the feature representation for a segment of sensor data, it becomes essential to devise a method to process both the input textual prompt and the sensor data to generate a response. To accomplish this, the LLaMA Large Language Model is employed as the language decoder \cite{touvron2023llama}. Furthermore, a bridging layer is incorporated to reconcile our sensor representation with the subsequent language decoder.

\subsubsection{Primer on Querying Transformer}
Prior to processing the encoded sensor representation with LLaMA, we utilize the Querying Transformer \cite{li2023blip2} to bridge the sensor encoder's outputs to LLaMA's embedding space. The Querying Transformer (Q-former) is a bottleneck architecture designed to distill textual information from the outputs of an upstream encoder. The Q-former is a transformer-based model that utilizes a set of learned queries as input and uses cross-attention to interact with the embeddings of the upstream encoder. Compared to traditional methods, such as employing a linear layer or feedforward network, the Q-former is able to generate any number of tokens for the downstream language decoder. Moreover, the use of attention and cross-attention allows the Q-former to learn and output complex interdependencies between the output tokens. As a result, the tokens produced are semantically coherent to the downstream LLM.

We initialize the Q-former with pre-trained weights to leverage the existing vision-language knowledge inherited from pre-training tasks \cite{li2023blip2} and further end-to-end text generation finetuning. Specifically, the pre-trained Q-former from VideoLLaMA's \cite{zhang2023videollama} omni-modal (AL) branch is utilized because it has undergone the aforementioned pre-training as well as further finetuning on vision-language tasks using ImageBind as the visual encoder, serving as the ideal base model to transfer knowledge to sensor data.

\subsubsection{Language Decoder}
The language decoder not only interprets data from the sensor encoder but also handles conversational tasks involving reasoning, leveraging world knowledge, and following instructions. Hence, we opt for a pre-trained Large Language Model (LLaMA-7B) \cite{touvron2023llama} as our optimal choice. LLaMA-7B, the most lightweight model in the LLaMA family, is well-suited for our computationally intensive training process.

Similar to other transformer-based large language models \cite{openai2024gpt4, anthropic2024claude}, LLaMA employs a text tokenizer to generate embeddings from an input text sequence, represented as $\text{Tokenizer}(x) \in \mathbb{R}^{T \times d_\text{LLaMA}}$, before processing it in the transformer layers. Here, $T$ is the number of tokens in the text sequence and $d_\text{LLaMA}$ indicates the size of each token embedding. The output of the language model can then be written as $y = \text{LLaMA}(\text{Tokenizer}(x))$.

To incorporate the sensor data, we apply the Querying Transformer to the output of the sensor encoder to extract sensor embeddings and prepend them to the text embeddings. The resulting output is $y = LLaMA(QFormer(\textbf{u}_{sens}) \oplus Tokenizer(x))$. Here, $\textbf{u}_{sens}$ is the output from the sensor encoder, and $QFormer(\textbf{u}_{sens}) \in \mathbb{R}^{K \times d_\text{LLaMA}}$ is the output from the Q-former, where $K$ denotes the number of learned queries in the Q-former and $d_\text{LLaMA}$ denotes the size of LLaMA's embedding space. Together, the computed sensor tokens and input text tokens allow the model to respond conditioned on sensor inputs.

\subsubsection{Temporal-Aware Tokens}
The sensor representation encapsulates information regarding events within a sequence of sensor data but lacks temporal clarity for the LLM. However, comprehending daily activities necessitates differentiating the chronological sequence of events, which may be further inquired by the user in detail. To tackle this issue, $n$ segments of equal length are sampled from a specified time interval of wearable sensor data. Each segment is then encoded to yield $\textbf{u}_{sens, t}$ for $t \in [1, ..., n]$. These encoded segments are concatenated with the output tokens from all $n$ intervals when inputted into LLaMA. The final output is represented as $y = LLaMA(QFormer(\textbf{u}_{sens,1}) \oplus ... \oplus QFormer(\textbf{u}_{sens,n}) \oplus Embedding(x))$. Given that each of the $n$ sequences of tokens carries meaningful information, different values for $n$ can be employed throughout training and inference. This adaptability enables Sensor2Text to handle variable lengths of wearable sensor data sequences.

Finally, the outputs of all of the $QFormer(\textbf{u}_{sens,i})$ are encapsulated in natural language, enabling us to leverage more of LLaMA's inherent capabilities. An illustrative prompt is shown in Figure \ref{fig:llm_prompt}.

\begin{figure}[h]
{
\fontsize{9.5pt}{11pt}\selectfont
\centering

\centering
\begin{tabular}{>{\centering\arraybackslash}m{2cm}m{12cm}}
\hline
\textbf{Stage} & \textbf{Prompt / Ground Truth} \\
\hline
Cross-Modal Training & \textbf{LLM Input:} \newline \verb|[INST]| \verb|<<SYS>>| You are a helpful language and vision assistant. You are able to understand the visual content that the user provides, and assist the user with a variety of tasks using natural language. \verb|<</SYS>>| \newline Open your eyes and imagine you see: \tcbox{\textless sensor embeddings\textgreater}. Provide a brief description of the scene. \verb|[/INST]| \newline \textbf{Ground Truth:} \newline A person is slicing a cucumber. \\
\hline
Instruct Finetuning & \textbf{LLM Input:} \newline \verb|[INST]| \verb|<<SYS>>| You are a helpful language and vision assistant. You are able to understand the visual content that the user provides, and assist the user with a variety of tasks using natural language. \verb|<</SYS>>| \newline \tcbox{\textless image embeddings\textgreater} What objects are the main focus of the image? \verb|[/INST]| \newline \textbf{Ground Truth:} \newline The main focus of the image is a collection of stuffed toy bears. \\
\hline
\end{tabular}

}
\caption{Example prompts and ground truths for the two aforementioned finetuning stages. After the text is tokenized into textual embeddings, the image or sensor embeddings outputted by the Querying Transformer are inserted at the indicated location, allowing the language model to process both the text and the image or sensor information when producing a response. For stage 1, using our hyperparameters, there are 64 sensor embeddings and approximately 100 text embeddings (depending on the training prompt). The same system prompt is used as the pre-trained vision model to maximize the effectiveness of transfer learning.}
\label{fig:llm_prompt}
\end{figure}

\subsubsection{Cross-Modal Training}

The language decoder is trained by employing a text generation loss over sensor/caption pairs. This loss is computed utilizing the logits generated by LLaMA for each token in its vocabulary and calculating a cross-entropy loss against the ground truth token. The total loss across a sequence of input tokens $\{x_i\}_{i=1,...,n}$ and output tokens $\{y_i\}_{i=1,...,m}$ is the cumulative sum of cross-entropy losses for each output token, conditioned on the preceding sequence of tokens. These tokens comprise a prompt, as illustrated in Figure \ref{fig:llm_prompt}, and the outputs of the Q-former conditioned on wearable sensor data. The total loss is expressed as follows:
\[
\mathcal{L}_{\text{gen}}(x, y; \theta) = -\sum_{i=1}^{m} \log p_\theta(y_i | x, y_1, ..., y_{i-1})
\]

Here, $\theta$ denotes the weights of the language decoder and $p_\theta$ is the probability that LLaMA assigns a given token. The loss is efficiently computed using a causal attention mask. This mechanism ensures that the model calculates output logits for a token-based solely on preceding tokens, allowing simultaneous computation of all summands in the generation loss for a given input/output sequence.

During training, LLaMA is frozen to preserve its language modeling abilities. In addition, the sensor encoder is frozen to promote training stability and uphold alignment between the embedding space of the sensor encoder and ImageBind's visual encoder. The loss incurred in text generation is transmitted through LLaMA to the Q-former via the tokens it generates in the input token sequence $\{x_n\}$.

Lastly, we rephrase activity labels into full, natural-language captions. This approach effectively leverages LLaMA's existing textual reasoning abilities, avoiding a substantial decline in language performance that naive fine-tuning might cause. The Q-former is trained using text generation loss with the modified input/output pairs and subsequently learns to produce tokens meaningful to LLaMA and prompt accurate generation outputs after the training.

\subsubsection{Noise Injection for Robustness}

Visual LLMs trained in a similar fashion \cite{li2023blip2, zhang2023videollama, liu2023visual} utilize much larger datasets than existing wearable sensor datasets. Consequently, the language decoder may be less robust, susceptible to overfitting, and exhibit significantly different behaviors with minor variations in the input distribution. To mitigate this issue,  noise injection is incorporated into the outputs of the Q-former before feeding them into LLaMA. Noise injection involves adding a random Gaussian vector with a small variance to the inputs, enhancing the model's resilience to minor input alterations. Specifically, we implement
\[
QFormer'(\textbf{u}_{sens})_i = QFormer(\textbf{u}_{sens})_i + \mathbf{X}_i, \mathbf{X}_i \sim \mathcal{N}(\mathbf{0}, \sigma^2 I_n)
\]

where  $QFormer(\textbf{u}_{sens})_i$ denotes the i-th token generated by the Q-former,  $\sigma^2$ represents a small variance, and $I_n$ is an identity matrix with n rows and n columns. Since adjacent points in LLaMA's embedding space share similar semantic meanings, this approach ensures robustness without substantially altering LLaMA's perception of its inputs in the training stage. Following the incorporation of noise injection, our model demonstrates improved performance on unseen data, as detailed in subsequent sections.

\subsubsection{Instruct Finetuning}
After completing the initial stage of cross-modal training, the model's performance in conversational contexts, such as responding to user inquiries, diminishes, as it has been trained solely to output activity labels. To address this issue, a second stage of fine-tuning is introduced to restore instruction-following capabilities. In this phase, the model is finetuned using question-answer pairs and example conversations. However, due to the lack of such a dataset with wearable sensor data, a visual instruction dataset is utilized instead. Since the sensor encoder has been trained with the ImageBind visual encoder as a teacher, the language decoder can be finetuned using ImageBind as the encoder on a vision instruct dataset. The Q-former is fine-tuned using text generation loss on image/conversation pairs, using ImageBind to extract visual embeddings. Through this process, the model retains its understanding of sensor data while also regaining instruction-following capabilities. After training is completed, the model can engage in conversations about sensor data without prior exposure to sensor-based discussions.

\subsection{Training Procedure and Implementation Details}
We select the datasets ActionSense \cite{delpretoLiu2022actionSense} and MMAct \cite{kong2019mmact} to evaluate our model. Each consists of subjects performing various daily activities while collecting wearable sensor data, corresponding video footage, and textual activity labels; further details regarding each dataset are elucidated in the following section. We sample the sensor data using a constant sampling rate of 50Hz or 100Hz, depending on the sensor data modality. We further apply preprocessing steps, such as filtering and normalization, to each wearable sensor data modality, as specified in the following section. Each video is cropped to 224 by 224 pixels and values across each color channel are normalized. These steps ensure that negative effects from faulty sensor readings or video footage are minimized.

We select paired sensor and video clips with a length of 2 seconds to train the sensor encoder. We set the size and stride of each 1D convolutional tokenizer in the model to 10. The sensor encoders for each modality consist of 6 transformer encoder layers, each with a hidden size of 512. Additionally, they feature a feedforward network with a hidden layer size of 2048 and multi-head attention with 8 attention heads and a dimension of 64 per head. Layer Normalization \cite{ba2016layer} is applied after each attention and feedforward layer in the transformers. Dropout regularization with a dropout rate of 0.1 is also employed. The encoder is trained using the Adam optimizer \cite{kingma2017adam} with a learning rate of $2*10^{-4}$ and a batch size of 32. Training is conducted by using the auxiliary training objective until validation loss stabilizes.

For the language decoder, a Q-former is employed, comprising $K=8$ query tokens, each with a dimension equivalent to LLaMA-7B's embedding dimension of 4096. $n=8$ temporal tokens are selected, and noise injection is applied with a variance of $\sigma^2 = 10^{-4}$. In the first stage of fine-tuning, the Q-former undergoes fine-tuning using the Adam optimizer \cite{kingma2017adam} with a learning rate of $2*10^{-6}$ on an NVIDIA T4 GPU for 20 epochs, during which the validation loss stabilizes. Moving to the second stage of fine-tuning, the LLaVA-Instruct dataset \cite{liu2023visual} is chosen, containing images from the Microsoft COCO \cite{chen2015microsoft} dataset alongside corresponding example conversations. ImageBind is utilized to generate visual embeddings for images in LLaVA-Instruct. Training in this stage follows the same settings as the first one, with the learning rate adjusted to $3*10^{-6}$ and conducted for 5000 iterations, leading to a qualitative assessment indicating the restoration of instructive capabilities.
\section{Evaluation}
In this section, a comprehensive evaluation of Sensor2Text's capabilities is provided. We showcase sample conversations with Sensor2Text, demonstrating its proficiency in daily activity captioning, question answering, and reasoning. We also systematically compare Sensor2Text against similar models using quantitative metrics. Additionally, we explore Sensor2Text's proficiency in leveraging multiple modalities of sensor data, its robust ability to generalize to unseen users, and ablation studies to assess the contribution of various components of the model. Finally, we assess Sensor2Text's performance when applied on low information-density IMU sensors.

\subsection{Dataset}
\begin{figure}
\centering
\includegraphics[width=1\textwidth]{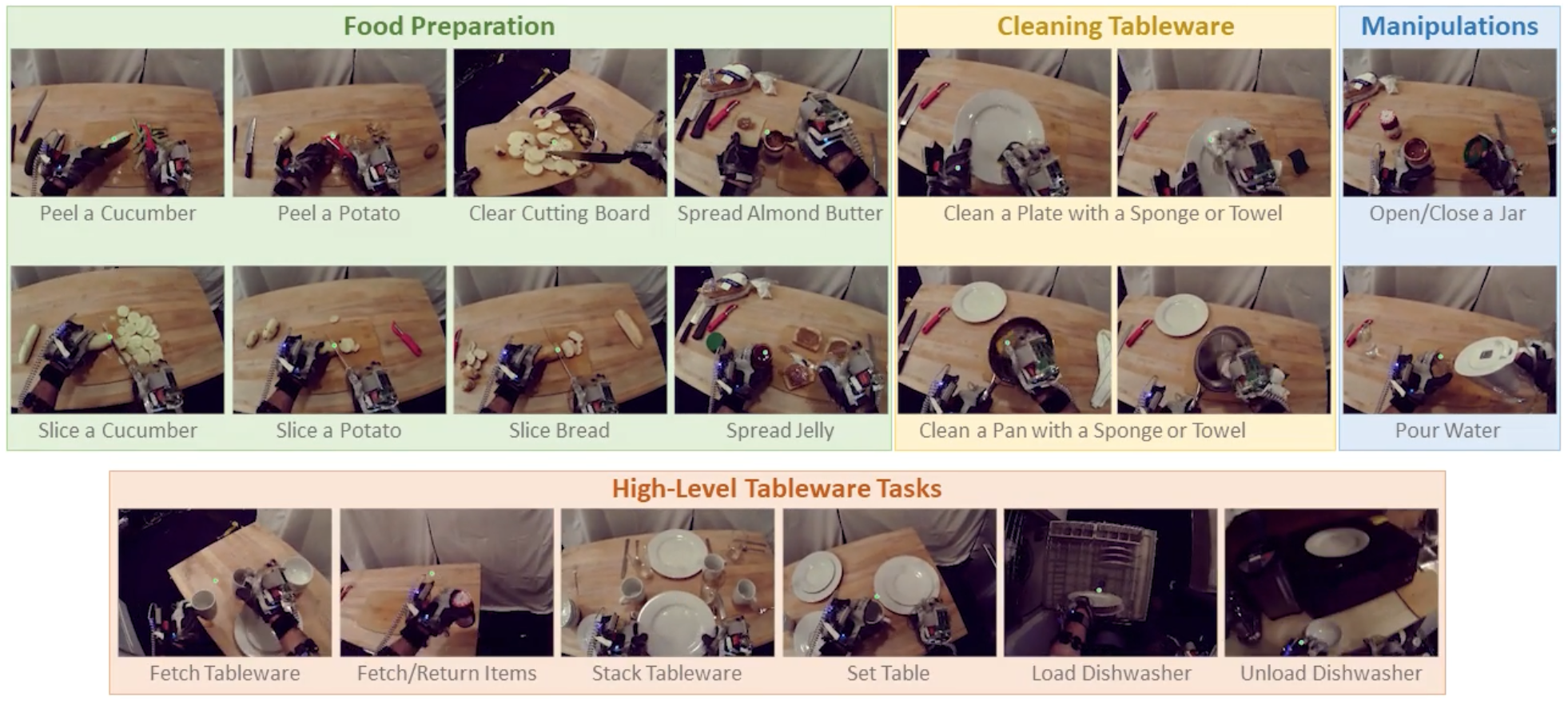}
\caption{ActionSense dataset. It consists of a variety of kitchen tasks, each with corresponding wearable sensor data, video data, and activity labels. Subjects perform a subset of these tasks over episodes of approximately 1 hour, with 10 total subjects.}
\label{fig:actionsense}
\end{figure}

We select the ActionSense \cite{delpretoLiu2022actionSense} dataset, a multi-modal wearable sensor dataset for human activities in a kitchen environment, for use in the primary qualitative and quantitative evaluations of the model. The ActionSense dataset consists of 10 subjects, each performing kitchen activities in episodes spanning 60-120 minutes, such as preparing food, cleaning tableware, and interacting with items. In total, there are 20 diverse kitchen activities distributed between these categories. The full set of activities is depicted in Figure \ref{fig:actionsense}.

Various wearable sensor data modalities are collected during each episode and are paired with corresponding visual footage and text labels. From the available data modalities, we select the body-tracking, muscle activation, and eye-tracking due to them being consistently present for all episodes. The data is gathered using the Xsens MTw Awinda body tracking system \cite{XsensMTwAwinda}, the Pupil Core eye-tracking headset from Pupil Labs \cite{PupilCore}, EMG sensors from Thalmic Lab's Myo Gesture Control Armband. Each wearable sensor records readings stored as sequential data of different frequencies. Additionally, we use visual footage collected from on-head cameras.


\subsubsection{Data Preprocessing}
To ensure the strong performance and reliability of our model, we preprocess and normalize each modality of wearable sensor data. We sample each at a frequency of 50Hz. For eye-tracking gaze data, we remove outliers by clipping values outside the range of $[0.05, 0.95]$, handle missing or faulty values through interpolation, and normalize the data to the range $[-1, 1]$. For Electromyography (EMG) data, we apply a low-pass filter with a cutoff frequency of 5Hz to smooth the signal and remove high-frequency noise, followed by normalizing each data value to the range $[-1, 1]$. Lastly, for the body tracking data, we preprocess the joint rotation values by normalizing them to the range $[-1, 1]$, assuming that the joint rotation data falls within the range of $[-180, 180]$ degrees. These preprocessing steps handle potential faulty sensor readings, thus enhancing the input data's quality and improving our deep learning architecture's performance.

\begin{figure}[h]
    {
    \fontsize{9.5pt}{11pt}\selectfont
    \begin{tabular}{p{3.5cm} p{1.5cm} p{9cm}}
        \toprule
        \multicolumn{3}{l}{\textbf{Sensor and Vision LLM Comparison}} \\
        \hline
        \\[-0.9em]
        \multirow{3}{*}{\includegraphics[width=1\linewidth]{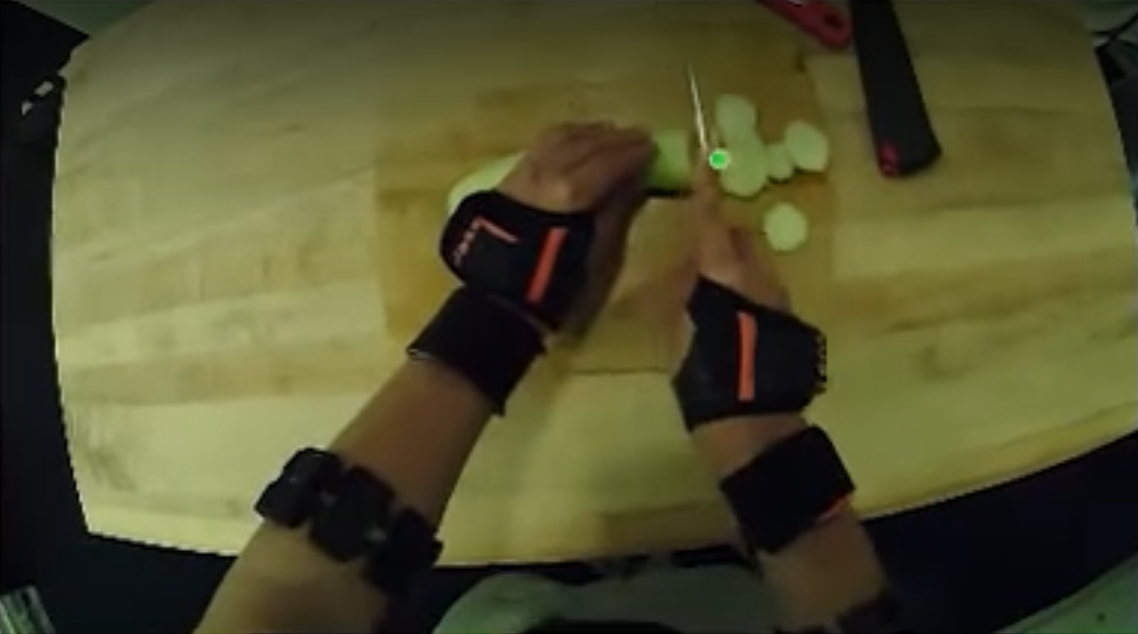}} & User & Describe what is happening in this scene. \\
        & VideoLLaMA & The person is cutting a cucumber. \\
        & User & Do they seem like they're good at cooking? \\
        & VideoLLaMA & They seem to be very skilled at slicing the cucumber. \\
        \cline{2-3}
        \\[-0.9em]
        & User & Describe what is happening in this scene. \\
        & GPT4 & The image shows a person's forearms, wearing wrist guards with orange detailing, cutting a small, light-colored object on a wooden cutting board with a knife. The setting appears to be from a first-person perspective, possibly from a head-mounted camera. \\
        & User & Do they seem like they're good at cooking? \\
        & GPT4 & It’s not clear from this one action whether they’re good at cooking; we’d need to see more of their technique and the final dish to judge their cooking skills! \\
        \hline
        \hline
        \\[-0.9em]
        \multirow{3}{*}{\includegraphics[width=1\linewidth]{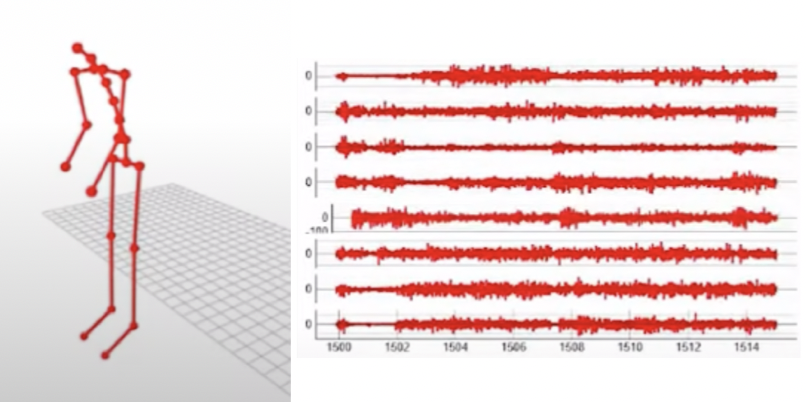}} & User & Describe what is happening in this scene. \\
        & Sensor2Text & A person is slicing a cucumber. \\
        & User & Do they seem like they're good at cooking? \\
        & Sensor2Text & The person seems to be very skilled at slicing the cucumber, as they are doing it with a lot of precision and care. \\
        \bottomrule
    \end{tabular}
    }

\caption{Comparison between Sensor2Text, GPT4, and fine-tuned VideoLLaMA. Each model demonstrates similar chat capabilities. However, Sensor2Text recalls the details of the subject's actions more precisely, demonstrating the advantages of using wearable sensors over video for some activities.}
\label{fig:example-comparison-good}
\end{figure}

We partition the dataset into train, validation, and test subsets following a 70/15/15 split ratio, respectively. The episodes in ActionSense are segmented into 2-second clips to train the sensor encoder, which significantly increases the number of training examples available. For the language decoder, labeled sections of each episode are chosen, segmented into 16-second clips, and then trained using text generation loss. In total, there are 16000 2-second segments for the cross-modality supervised training of the sensor encoder and 1100 16-second segments that are adequately labeled. After incorporating a curated list of training prompts for the latter, there are approximately 7700 training examples for end-to-end text generation training.

\subsection{Qualitative Evaluation}

This section presents examples of Sensor2Text in diverse conversational contexts. For each instance, wearable sensor data from an unseen subject in the ActionSense dataset is utilized, and multiple questions are posed to the chatbot to assess its responses. For qualitative analysis, ground truth of the tested human activities is illustrated using footage from video cameras. These recordings are subsequently inputted into vision-language models to facilitate comparison with other models of the same category. It's worth noting that these visual data were not provided to Sensor2Text; the sensor-language model only receives the aforementioned wearable sensor data modalities as input. The following capabilities of Sensor2Text were evaluated and will be further discussed in the following subsections: sensor perception ability, conversational capabilities, and external knowledge integration. Additionally, a qualitative comparison between Sensor2Text and existing video language models is provided, including GPT4 \cite{openai2024gpt4}, Claude Opus \cite{anthropic2024claude}, and a fine-tuned open source model VideoLLaMA \cite{zhang2023videollama}. The performance of the sensor-language model is also compared against vision models using footage from inadequate lighting or unfavorable viewpoints, showing its superior performance in these settings.

\begin{figure}[h]
    {
    \fontsize{9.5pt}{11pt}\selectfont
        \begin{tabular}{p{3.5cm} p{1.7cm} p{9cm}}
        \toprule
        \multicolumn{3}{l}{\textbf{Sensor and Vision LLM Comparison, Poor Lighting}} \\
        \hline
        \\[-0.9em]
        \multirow{3}{*}{\includegraphics[width=1\linewidth]{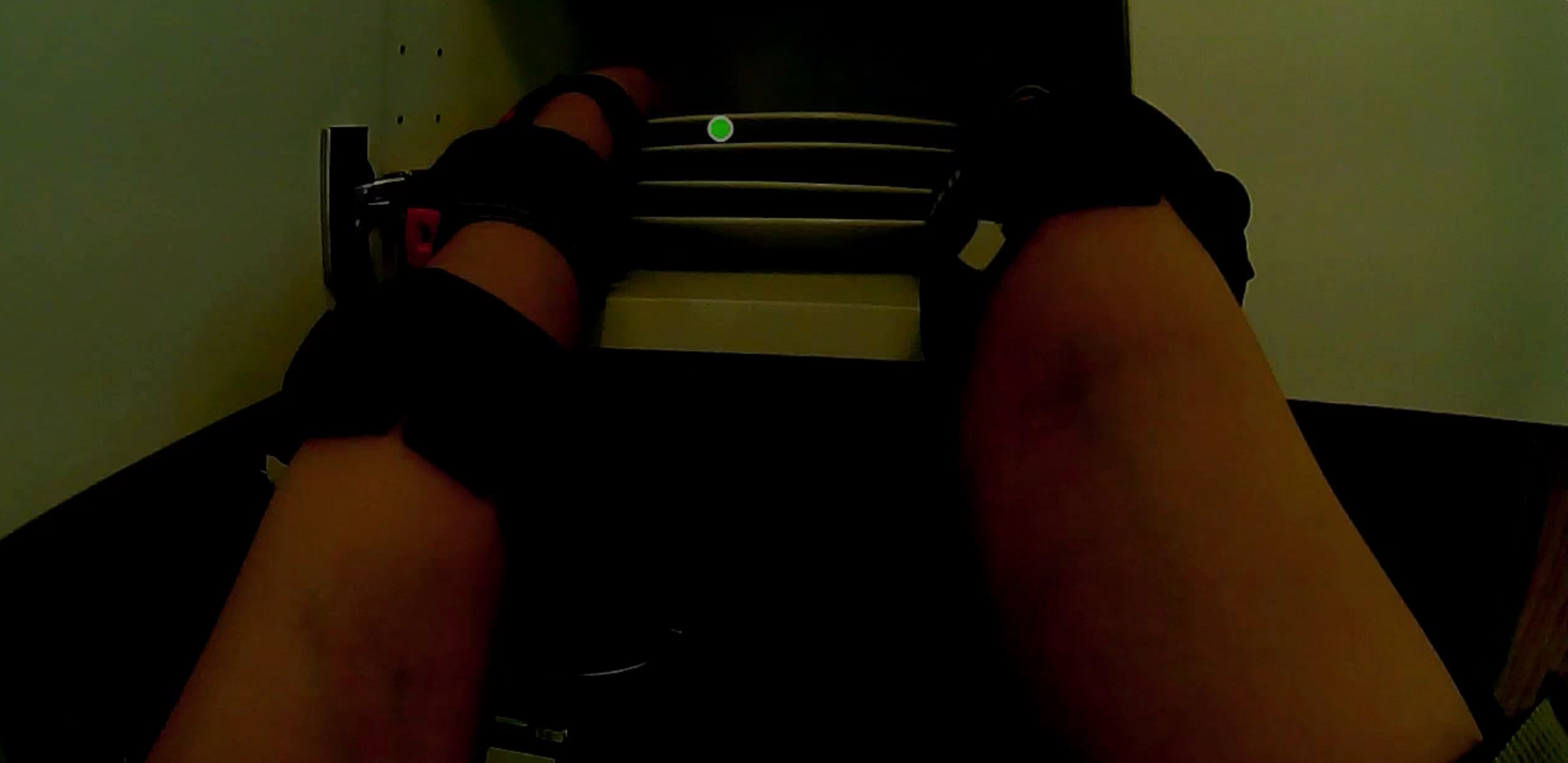}} & User & Describe what is happening in this scene. \\
        & VideoLLaMA & The scene is a man sitting on a couch, looking at a computer screen. \textcolor{red}{(incorrect)} \\
        \cline{2-3}
        \\[-0.9em]
        & GPT-4 & This image shows someone’s knees bent at what appears to be a close-up angle, with the person wearing knee braces or supports. It seems like they are undergoing some sort of physical therapy or medical treatment, possibly using a machine designed for knee or leg exercises. The lighting is quite low, so the details are not very clear. \textcolor{red}{(incorrect)} \\
        \cline{2-3}
        \\[-0.9em]
        & Claude Opus & The image shows a person's legs and feet resting on what appears to be a desk or table. The room is dimly lit, with a green light visible near the top of the frame, possibly coming from a device like a modem or router. The person is wearing black shoes or sandals. Without being able to see more of the scene or identify the individual, there's not much additional context I can provide about what exactly is happening, but it seems to depict someone sitting at a desk or workstation in a dimly lit room or office. \textcolor{red}{(incorrect)} \\
        \hline
        \hline
        \\[-0.9em]
        \multirow{3}{*}{\includegraphics[width=1\linewidth]{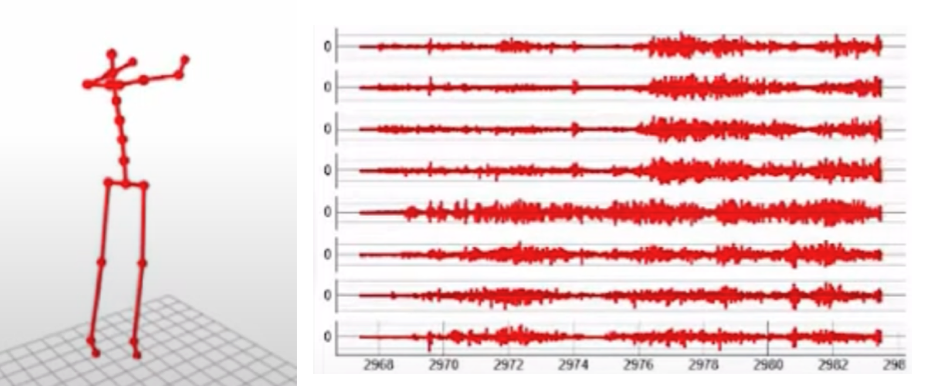}} & User & Describe what is happening in this scene. \\
        & Sensor2Text & A person is \textcolor{blue}{getting items from a cabinet}. \\
        & User & Is the cabinet above or below the person? \\
        & Sensor2Text & The cabinet is \textcolor{blue}{above} the person. \\
        & & \textcolor{blue}{(correct)} \\
        
        \bottomrule
    \end{tabular}
    }

\caption{Comparison between Sensor2Text and vision-based models on dark examples. The sensor-based model correctly determines the subject's actions and the locations of objects in the scene. In contrast, the vision-based models are unable to determine the subject's activity.}
\label{fig:example-comparison-bad}
\end{figure}

\begin{figure}[h]
    {
    \fontsize{9.5pt}{11pt}\selectfont
        \begin{tabular}{p{3.5cm} p{1.7cm} p{9cm}}
        \toprule
        \multicolumn{3}{l}{\textbf{Sensor and Vision LLM Comparison, Unfavorable Viewpoints}} \\
        \hline
        \\[-0.9em]
        \multirow{3}{*}{\includegraphics[width=1\linewidth]{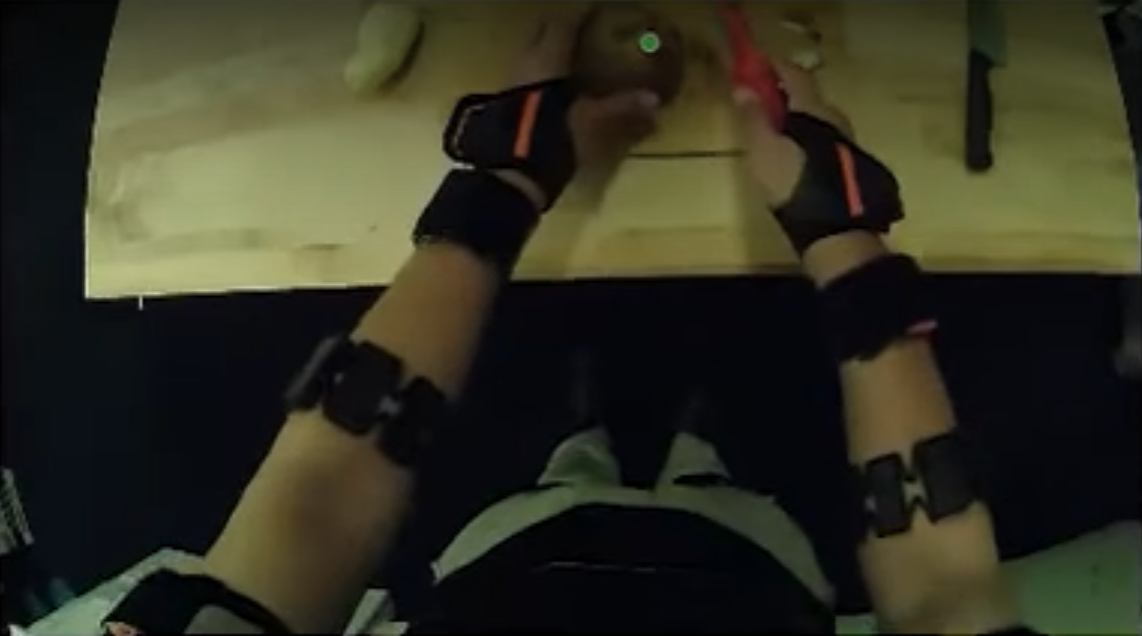}} & User & What is the person doing? \\
        & VideoLLaMA & The person is cutting a cake with a knife. \textcolor{red}{(incorrect)} \\
        \cline{2-3}
        \\[-0.9em]
        & GPT-4 & The person in the image appears to be in the middle of a physical activity, possibly exercising or participating in some kind of sport. They are wearing what looks like athletic gear, including gloves and possibly wrist guards or fitness trackers on both arms. Due to the perspective of the image, which seems to be taken from a camera mounted on the person's head or chest, it's difficult to provide precise details, but it's clear that the person is engaged in a physical, probably sports-related, activity. \textcolor{red}{(incorrect)} \\
        \cline{2-3}
        \\[-0.9em]
        & Claude Opus & The image shows a person lying down, likely on a bed or couch. Their legs and feet are visible, wearing what appear to be black knee pads or braces. Without seeing more of the person or surrounding context, it's difficult to say exactly what they are doing, but they seem to be resting or relaxing based on their reclining position. \textcolor{red}{(incorrect)} \\
        \hline
        \hline
        \\[-0.9em]
        \multirow{3}{*}{\includegraphics[width=1\linewidth]{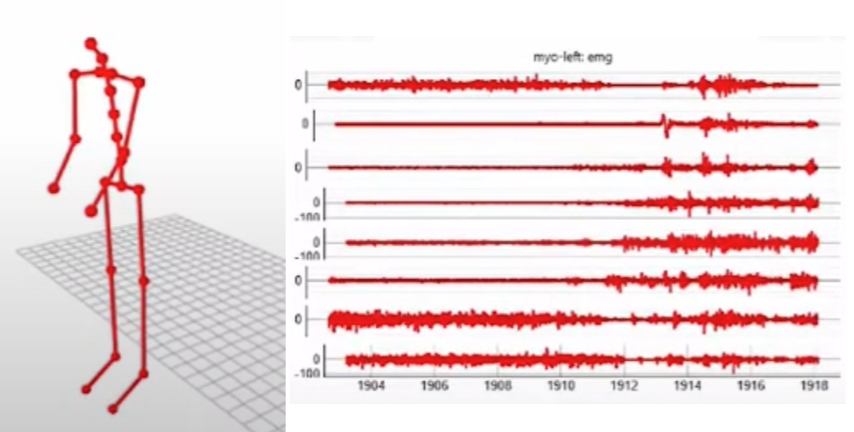}} & User & What is the person doing? \\
        & Sensor2Text & A person is peeling a potato. \\
        & User & How long do you think this will take? \\
        & Sensor2Text & It will take about 30 seconds. \\
        & & \textcolor{blue}{(correct)} \\
        
        \bottomrule
    \end{tabular}
    }

\caption{Comparison between Sensor2Text and vision-based models on an activity with a bad on-head camera angle. Again, Sensor2Text performs equally well and correctly determines the activity, while all of the vision-based models hallucinate details and are unable to answer correctly.}
\label{fig:bad-viewpoint-comparison}
\end{figure}

\subsubsection{Sensor Perception Ability}
Figure \ref{fig:example-comparison-good} illustrates Sensor2Text's proficiency in identifying various activities from sensor data. Sensor2Text correctly identifies that the subject is slicing a cucumber and notes the subject's precise movements during the slicing process, showing its powerful sensor perception abilities.

Furthermore, in Figure \ref{fig:example-comparison-bad}, Sensor2Text accurately identifies that the subject is retrieving items from a cabinet and correctly identifies that the cabinet is above the subject. In contrast, all three video models are unable to determine any details of the activities due to the inadequate lighting. In Figure \ref{fig:bad-viewpoint-comparison}, during intervals with bad camera angles, the sensor model is again able to determine the subject's activity while vision models fail. This shows the superior performance of the sensor-based model in diverse settings.

\subsubsection{Conversational Capability}
Sensor2Text excels at interactive conversations with users, seamlessly blending the conversational abilities of LLaMA with insights derived from provided sensor data. Illustrated in Figure \ref{fig:example-comparison-good}, when prompted, Sensor2Text correctly identifies that the subject is slicing a cucumber. When asked if the subject appears good at cooking, the model responds affirmatively and cites the subject's precise movements during the slicing process. This demonstrates Sensor2Text's ability to integrate the information from the inputted sensor data together with the language model's reasoning capabilities for smooth conversations. This integration enables dynamic and contextually aware interactions.

\subsubsection{External Knowledge Integration}
Sensor2Text is capable of drawing upon world knowledge while integrating insights from provided sensor data. In Figure \ref{fig:bad-viewpoint-comparison}, Sensor2Text correctly assesses that it will take the subject about 30 seconds to peel the potato, showcasing its ability to utilize the information provided by the sensors and its inherent world knowledge. This enables dynamic and context-aware interactions.

\subsection{Quantitative Evaluation}
\subsubsection{Evaluation Metrics}
Providing a quantitative evaluation for Sensor2Text is challenging due to the unstructured nature of conversation-based tasks. While visual language models have established benchmarks like VQA \cite{VQA}, LLaVA-Bench \cite{liu2023visual}, and ScienceQA \cite{lu2022learn}, which assess conversational models' performance via meticulously crafted tasks, prompts, and metrics, comparable benchmarks are lacking for sensor language tasks. For example, Visual Question-Answering (VQA) assesses a model's correctness by comparing its output against a large set of human annotations \cite{VQA}. In the absence of such benchmarks for sensor language tasks, we employ captioning scores as a proxy for evaluating performance, leveraging the availability of ground truth captions in our training data.

To assess Sensor2Text's performance, the model is prompted with a generic prompt requesting a caption. Such a prompt is illustrated in Figure \ref{fig:llm_prompt}. Then, traditional image captioning benchmarks are used to score the model's response. Following Microsoft COCO \cite{chen2015microsoft}, we utilize the BLEU \cite{Papineni}, ROUGE-L \cite{lin-2004-rouge}, METEOR \cite{banerjee-lavie-2005-meteor}, CIDEr \cite{vedantam2015cider}, and SPICE \cite{anderson2016spice} evaluation metrics. BLEU considers the co-occurrences of words between the predicted and ground truth sentences, with BLEU-n accounting for the co-occurrences of n words. ROUGE-L measures the longest common subsequence between the predicted and ground truth captions. METEOR computes the alignment between captions by considering exact token matches, stemming, and paraphrase matching, taking into account precision, recall, and word order. CIDEr assesses the similarity between a generated caption and a set of ground truth captions using TF-IDF weighting for each n-gram. SPICE evaluates the semantic similarity by extracting semantic propositions from the captions and computing the F1 score between the propositions of the predicted and ground truth captions. Together, these scores provide a comprehensive evaluation of the captioning performance of language models.

\subsubsection{Captioning}
The LLMs are prompted with a randomly selected prompt from a predefined list to elicit a caption output. This approach encourages the language model to produce outputs closely aligned with the terse, one-sentence descriptions of scenes found in the ActionSense dataset annotations.

Table \ref{tab:eval_scores} showcases the quantitative captioning results where the model undergoes evaluation against various state-of-the-art vision-based Language Models (LLMs). Specifically, a comparison is drawn with VideoLLaMA's Audio-Language (AL) branch \cite{zhang2023videollama}, representing the omni-modal (video and audio) branch of VideoLLaMA and the source for the pre-trained Q-former fine-tuned for Sensor2Text. Additionally, comparisons are made against existing state-of-the-art closed-source models, GPT4 \cite{openai2024gpt4} and Claude Opus \cite{anthropic2024claude}. VideoLLaMA receives prompts based on the entire video, whereas GPT4 and Claude Opus are prompted with a representative frame sampled from the video. Finally, we provide a comparison against a powerful visual captioning model, UniVL \cite{luo2020univl}, which serves as a benchmark for video captioning.

To ensure a fair comparison between Sensor2Text and the selected vision-LLM models, additional techniques are employed to optimize their performance. For the open-source model VideoLLaMA, we apply additional fine-tuning by freezing the weights of the visual encoder and language decoder, and subsequently training the Q-former using text generation loss. This process follows a methodology similar to VideoLLaMA's original training procedure. The captioning scores for Sensor2Text and the fine-tuned VideoLLaMA are reported after stage 1 of finetuning, where the models are primed to generate the most direct and caption-like responses.

For the closed-source models, GPT4 and Claude Opus, a list of all activity labels is included in the prompt alongside the visual input and the LLM is repeatedly prompted until its response matches one of the provided labels. This partially compensates for the models' lack of finetuning on the dataset. Additionally, we provide captioning scores for VideoLLaMA using this prompting method without finetuning. All models are evaluated on the same test set as Sensor2Text, ensuring a consistent evaluation process.

Furthermore, benchmarking is conducted against existing video captioning models as well, specifically UniVL \cite{luo2020univl}, after being fine-tuned on our dataset to facilitate an accurate comparison. It's important to note that multi-modal LLM models are disadvantaged when benchmarked against video-captioning models using captioning scores because captioning models are tuned to this specific task while multi-modal LLMs have powerful capabilities outside of captioning.

\begin{table}[h]
\centering
\begin{tabular}{l|c|c|c|c|c|c|c|c}
\hline
Method & B@1 & B@2 & B@3 & B@4 & R & M & C & S \\
\hline
VideoLLaMA & 41.0 & 35.4 & 29.7 & 22.4 & 42.0 & 19.9 & 8.7 & 23.6 \\
Claude Opus & 64.3 & 57.7 & 51.6 & 44.9 & 65.7 & 33.7 & 33.9 & 48.8 \\
GPT4 & 72.3 & 67.8 & 63.8 & 59.4 & 75.7 & 42.3 & 54.0 & 64.6 \\
VideoLLaMA w/ finetuning & 80.6 & 76.5 & 73.5 & 70.4 & 81.4 & 48.7 & 66.2 & 73.0 \\
UniVL w/ finetuning & 82.6 & 78.5 & 74.8 & 70.9 & 83.1 & 49.0 & 62.5 & 67.5 \\
\hline
Sensor2Text & 78.7 & 74.5 & 71.1 & 67.5 & 80.5 & 46.7 & 62.2 & 68.6 \\
\hline
\end{tabular}
\caption{Captioning scores on the ActionSense dataset. We denote BLEU-n as B@n, ROUGE-L as R, METEOR as M, CIDEr as C, and SPICE as S. All scores are normalized for a full score of 100. Sensor2Text's performance is superior to un-finetuned vision LLM models and similar to that of finetuned vision models.}
\label{tab:eval_scores}
\end{table}

These scores illustrate how Sensor2Text achieves similar performance compared to both finetuned visual LLMs and finetuned video captioning models when generating captions on the ActionSense dataset and outperforms well-prompted closed-source vision LLMs. Remarkably, even with a relatively small number of training examples, our training methodology effectively leverages knowledge from visual-language models and extends it to sensor data. Additionally, as illustrated in Figure \ref{fig:example-comparison-bad}, the sensor-based model is unaffected by poor lighting conditions, in which vision-based models perform very poorly. These results validate the feasibility of using wearable sensors as an alternative to video for tracking daily activities.

\subsubsection{Generalization and Robustness}

This section showcases Sensor2Text's generalization capacity to unseen users. Due to the slight variations in data collected from each user, additional fine-tuning is often necessary for each new user. This process can impede the practical deployment of Sensor2Text, as collecting such data is time-consuming and arduous, presenting a challenge for new users.

A comparison is conducted between the model's performance when evaluated on data from users it has encountered before versus data from unseen users. Specifically, rather than splitting the train/validation/test sets by uniformly sampling across the entire dataset, partitioning by user is conducted instead. A subset of users is selected for exclusion, and the model is trained on the remaining dataset for 20 epochs, mirroring the original model's training duration. Captioning performance is then evaluated on the held-out users. The captioning evaluation metrics comparing performance on data from seen users and unseen users are presented in Table \ref{tab:unseen_subjects}.

\begin{table}[h]
\centering
\begin{tabular}{l|c|c|c|c|c|c|c|c}
\hline
Method & B@1 & B@2 & B@3 & B@4 & R & M & C & S \\
\hline
unseen users & 73.8 & 69.6 & 66.2 & 62.5 & 77.7 & 43.3 & 58.9 & 68.0 \\
seen users & 78.7 & 74.5 & 71.1 & 67.5 & 80.5 & 46.7 & 62.2 & 68.6 \\
\hline
\end{tabular}
\caption{Captioning performance on the ActionSense dataset comparing totally unseen users versus seen users. Unseen users see a slight drop in performance, demonstrating Sensor2Text's powerful generalization capabilities.}
\label{tab:unseen_subjects}
\end{table}

Table \ref{tab:unseen_subjects} illustrates Sensor2Text's remarkable generalization capabilities and robustness. It exhibits only a minor decline in performance when applied to data from new users. This minimal degradation highlights the effectiveness of employing teacher-student networks and transfer learning in enhancing our model's generalization performance.

\subsection{Multi-Modality}

One crucial aspect of Sensor2Text lies in its usage of wearable sensor data across various modalities. This section examines the impact of multiple modalities on the model's training, contrasting it with the use of a single modality, accessed by conducting experiments where only a subset of data modalities is utilized. Specifically, several new models are trained using solely eye tracking, muscle activation, or body tracking data, following the same training procedure. The results are presented in the Table \ref{tab:holdout_sensors}.
\begin{table}[h]
\centering
\begin{tabular}{l|c|c|c|c|c|c|c|c}
\hline
Method & B@1 & B@2 & B@3 & B@4 & R & M & C & S \\
\hline

eye only & 52.2 & 44.4 & 38.2 & 31.0 & 52.4 & 24.3 & 17.4 & 34.1 \\
muscle only & 66.3 & 60.9 & 56.2 & 51.3 & 71.1 & 36.0 & 45.3 & 57.4 \\
body only & 73.4 & 67.9 & 63.6 & 59.1 & 74.5 & 42.2 & 53.4 & 62.9 \\
eye, muscle, body & 78.7 & 74.5 & 71.1 & 67.5 & 80.5 & 46.7 & 62.2 & 68.6 \\
\hline
\end{tabular}

\caption{Captioning scores achieved on ActionSense using only a subset of available sensor data modalities. Combining all modalities yields the best performance.}
\label{tab:holdout_sensors}
\end{table}

As demonstrated in the table, Sensor2Text shows stronger performance when given more modalities of sensor data, arising from the model's enhanced capability of differentiating between similar activities and mitigating the impact of noise in individual sensors. Such findings underscore Sensor2Text's effectiveness in leveraging diverse modalities of data, affirming our notion that the multiple data modalities of wearable sensor data provide advantages not present in video data.

\subsection{Ablation Study}%

To show the necessity of the various components, experiments are conducted on the model's architecture and training process. Specifically, we demonstrate the need for the 2-step finetuning process, noise injection, and usage time-aware embeddings in the language decoder through ablation studies. The following alternatives are considered:

\begin{itemize}
    \item Omitting language decoder fine-tuning
    \item Using a single embedding instead of 8 time-ordered embeddings
    \item Omitting noise injection
\end{itemize}

Each alternative is compared against the model using the same captioning scores as before. Consistency is maintained by keeping the sensor encoder and train/validation/test split constant. For the experiments that require re-training the language decoder, the same training procedures and applicable hyperparameters are employed as the baseline training. The results are shown in Table \ref{tab:ablation}.

\begin{table}[h]
\centering
\begin{tabular}{l|c|c|c|c|c|c|c|c}
\hline
Method & B@1 & B@2 & B@3 & B@4 & R & M & C & S \\
\hline
w/o finetuning & 41.8 & 35.2 & 28.7 & 19.0 & 44.3 & 17.1 & 7.2 & 22.5 \\
w/o temporal embeddings & 75.1 & 70.6 & 67.2 & 63.9 & 78.8 & 42.9 & 62.2 & 68.0 \\
w/o noise & 76.6 & 71.7 & 67.9 & 63.8 & 77.6 & 45.8 & 56.9 & 66.2 \\
Sensor2Text & 78.7 & 74.5 & 71.1 & 67.5 & 80.5 & 46.7 & 62.2 & 68.6 \\
\hline
\end{tabular}

\caption{Captioning results on the ActionSense dataset after omitting various components of the architecture.}
\label{tab:ablation}
\end{table}

As depicted in the table, Sensor2Text consistently exhibits the highest captioning scores across all evaluation metrics in all ablation studies, highlighting the improved performance achieved through finetuning, noise injection, and the integration of temporal tokens. In summary, these findings underscore the essential role of these components in optimizing the performance of the end-to-end model.


\subsection{Low Information Density}

Although Sensor2Text shows strong performance on the ActionSense dataset, the high information-density sensors present in the dataset may be difficult to collect in practice for daily activity tracking. For instance, collecting full body-tracking data involves wearing bulky sensors attached to a full-body suit. Thus, further investigation is needed to assess the applicability of Sensor2Text to sensors with low information density, such as smartphone and smartwatch IMUs. In this section, we select another dataset, the MMAct dataset \cite{kong2019mmact}, and provide a qualitative and quantitative evaluation to assess whether Sensor2Text can perform accurate Q\&A and have meaningful conversations when applied to the MMAct dataset.

\subsubsection{MMAct Dataset}

To this end, we select the MMAct dataset \cite{9009579}. In this dataset, 20 subjects perform activities from a set of 37 activities such as sitting, jumping, and entering/exiting a room. Wearable sensor data is collected from a smartphone and smartwatch, and the corresponding video is captured via environment-mounted video cameras. The smartphone collects acceleration, gyroscope, and orientation data, while the smartwatch further collects a set of smartwatch acceleration data for 4 total sensor data modalities. In contrast to the wearable sensor data collected from the ActionSense dataset, the sensors used in MMAct are more feasible to collect during everyday life but have lower information density.

To preprocess the data, we sample each modality of sensor data at a frequency of 50Hz. Each acceleration, gyroscope, and orientation signal is given in the three orthogonal directions $(x,y,z)$. We begin by computing the average magnitude across the dataset for each sensor modality and normalizing each to have an average magnitude of 1. The collected data may be sensitive to sensor placement; thus, following \cite{9009579}, we further compute $R_i = \text{arcsin}\left(\frac{z_i}{\sqrt{x_i^2 + y_i^2 + z_i^2}}\right)$ for each data point and concatenate this to the original data. In total, each sensor data modality consists of 4 data values, significantly lower than the 66 data values from ActionSense's body tracking data or 16 from ActionSense's EMG data, signifying the lower information density. We use 60\% of the data for training, 20\% for validation, and 20\% for testing. The same training settings are applied as before, with for 200 epochs  for the sensor encode,  20 epochs for the language decoder trained, and 5000 iterations for instruction finetuning.

\subsubsection{Qualitative Evaluation}
This section presents examples of Sensor2Text applied to activities from the MMAct dataset. Again, a comparison is presented with existing vision language models, including state-of-the-art closed-source LLMs. Sensor2Text is provided all 4 modalities of wearable sensor data but not the visual data. To facilitate an efficient comparison, we further configure a system prompt for the vision-language model to help the model filter out details such as the labels on the walls and instead focus on the subject. This ensures that its response is concise and focuses on details useful to our evaluation of daily activity perception ability. We compare and discuss Sensor2Text's performance against the qualitative results from ActionSense.

\begin{figure}[h]
    {
    \fontsize{9.5pt}{11pt}\selectfont
        \begin{tabular}{p{3.5cm} p{1.7cm} p{9cm}}
        \toprule
        \multicolumn{3}{l}{\textbf{Sensor and Vision LLM Comparison, IMU sensors, high-level scene understanding}} \\
        \hline
        \\[-0.9em]
        \multirow{3}{*}{\includegraphics[width=1\linewidth]{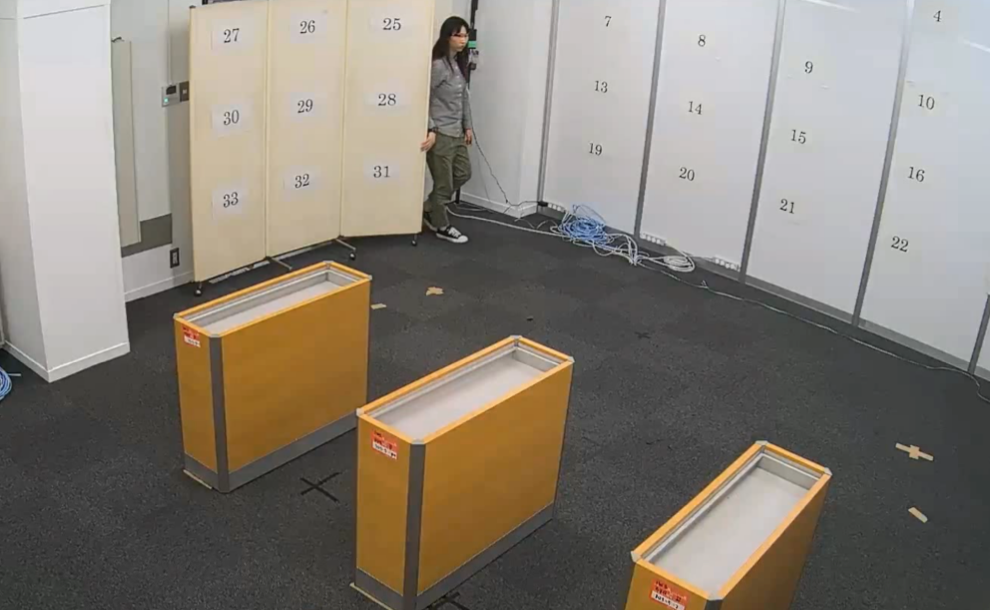}} & User & What is happening in this scene? \\
        & GPT-4 & The subject is standing near a wall, possibly adjusting or interacting with a device or panel on the wall. They appear to be stationary and focused on the task involving the wall-mounted object. \\
        & & \textcolor{red}{(incorrect)} \\
        & User & Does the person seem relaxed or in a hurry? \\
        & GPT-4 & The person appears relaxed, as there are no visible signs of haste or urgency in their posture or movements. \\
        & & \textcolor{blue}{(correct)} \\
        \hline
        \hline
        \\[-0.9em]
        \multirow{3}{*}{\includegraphics[width=1\linewidth]{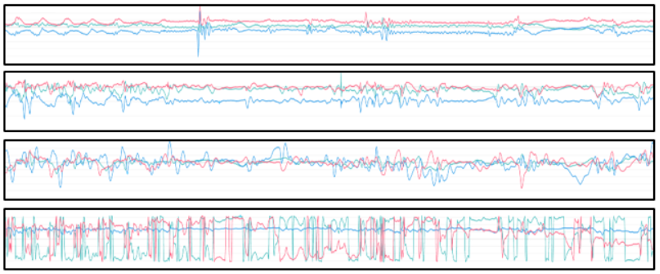}} & User & What is happening in this scene? \\
        & Sensor2Text & A person is entering the room. \\
        & & \textcolor{blue}{(correct)} \\
        & User & Does the person seem relaxed or in a hurry? \\
        & Sensor2Text & The person seems to be in a hurry. \\
        & & \textcolor{red}{(incorrect)} \\
        
        \bottomrule
    \end{tabular}
    }

\caption{Comparison between Sensor2Text and GPT-4 on the MMAct dataset, comparing both models' understanding of the scene. The subject cautiously enters the room from outside. Sensor2Text is able to determine the subject's activity from the movement data gathered by IMU sensors. However, it incorrectly answers that the person is in a hurry. In contrast, GPT-4 is unable to determine the subject's action because of its lack of temporal understanding but correctly answers that the person is relaxed.}
\label{fig:mmact-enter-comparison}
\end{figure}

\begin{figure}[h]
    {
    \fontsize{9.5pt}{11pt}\selectfont
        \begin{tabular}{p{3.5cm} p{1.7cm} p{9cm}}
        \toprule
        \multicolumn{3}{l}{\textbf{Sensor and Vision LLM Comparison, IMU sensors, detail recall}} \\
        \hline
        \\[-0.9em]
        \multirow{3}{*}{\includegraphics[width=1\linewidth]{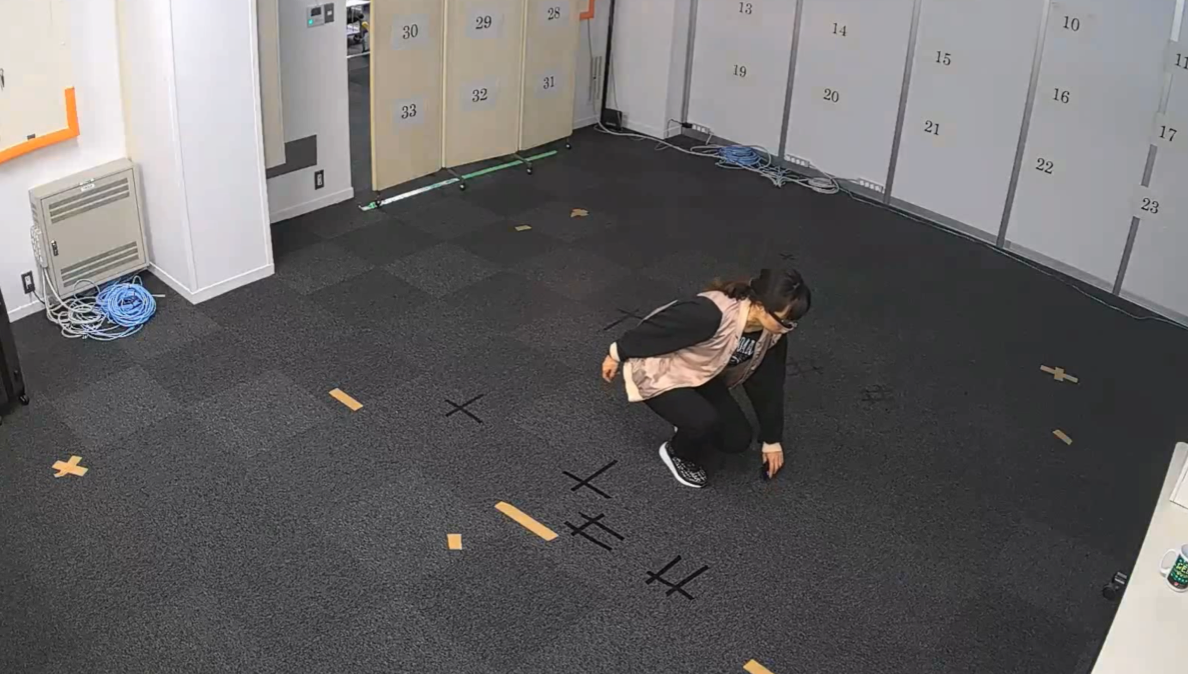}} & User & What is the person doing? \\
        & GPT-4 & The subject in the scene is crouching down and appears to be picking something up or placing something on the floor. \\
        & User & Describe the item they are picking up. What is its shape and size? Does it seem heavy or light? \\
        & GPT-4 & The item the subject is picking up is small and likely flat, given the way it is being handled. It seems to be light, as the subject is using only one hand to pick it up with ease. \\
        & & \textcolor{blue}{(correct)} \\
        \hline
        \hline
        \\[-0.9em]
        \multirow{3}{*}{\includegraphics[width=1\linewidth]{image/imu-sensors.png}} & User & What is the person doing? \\
        & Sensor2Text & A person is picking something up. \\
        & & \textcolor{blue}{(correct)} \\
        & User & Describe the item they are picking up. What is its shape and size? Does it seem heavy or light? \\
        & Sensor2Text & The item is a phone. It is light in weight and has a round shape. \\
        & & \textcolor{red}{(incorrect)} \\
        
        \bottomrule
    \end{tabular}
    }
\caption{Comparison of Sensor2Text and GPT-4 on the MMAct dataset, focusing on both models' ability to recall details. Both models are able to determine the subject's actions. GPT-4 is further able to estimate the properties of the object on the floor, whereas Sensor2Text hallucinates.}
\label{fig:mmact-pickup-comparison}
\end{figure}

As shown in Figures \ref{fig:mmact-enter-comparison} and \ref{fig:mmact-pickup-comparison}, Sensor2Text is able to leverage the smartphone/smartwatch sensors to correctly determine the activity being performed. In Figure \ref{fig:mmact-enter-comparison}, Sensor2Text correctly determines that the subject is entering the room, demonstrating its ability to function in low information density settings. However, Sensor2Text often fails to understand further details. In the same figure, when asked whether the subject appears to be in a hurry, Sensor2Text incorrectly replies yes, even though the subject is being notably cautious and moving slowly. In contrast, GPT-4 is unable to determine the subject's actions due to the lack of temporal context in images but is able to recognize that the subject is relaxed. In comparison to GPT-4, Sensor2Text applied on IMU sensors is more capable of daily activity tracking but less capable of grasping details. In Figure \ref{fig:mmact-pickup-comparison}, Sensor2Text is able to correctly determine that the subject is picking an item off the ground. However, when asked to describe details of the item, Sensor2Text is only able to correctly reply that the item is light but hallucinates other details about the item. This demonstrates that Sensor2Text is only able to use smartphone and smartwatch sensors to deduce a limited set of information about the activity due to the inherent limitations of the sensors themselves.

\subsubsection{Quantitative Evaluation}
A quantitative evaluation is further conducted on Sensor2Text, with captioning scores serving as a proxy to compare Sensor2Text's performance against existing vision language models. Following the same procedure as before, the list of possible captions to the prompt is prepended when prompting closed-source vision LLMs to ensure that their outputs are in the target distribution. The results are given in Table \ref{tab:mmact_eval_scores}. Again, Sensor2Text achieves higher captioning scores than vision LLM models without finetuning. This demonstrates Sensor2Text's remarkable ability to leverage low information-density sensors to determine the subject's activity. However, as demonstrated in the qualitative evaluation, Sensor2Text's performance in detail recognition decreases when applied to low information-density sensors. This shows that a decrease in the information density primarily causes a decrease in detail recall performance, whereas the model can still successfully leverage low information-density sensors to achieve activity recognition.

\begin{table}[h]
\centering
\begin{tabular}{l|c|c|c|c|c|c|c|c}
\hline
Method & B@1 & B@2 & B@3 & B@4 & R & M & C & S \\
\hline
VideoLLaMA & 52.5 & 46.2 & 38.3 & 19.5 & 56.2 & 30.2 & 2.8 & 31.9 \\
Claude Opus & 62.9 & 58.1 & 51.2 & 35.7 & 64.9 & 30.4 & 11.2 & 49.6 \\
GPT4 & 70.5 & 67.2 & 62.9 & 54.2 & 75.3 & 36.6 & 29.6 & 61.7 \\
VideoLLaMA w/ finetuning & 78.0 & 74.0 & 69.4 & 61.8 & 79.2 & 42.9 & 45.1 & 70.4 \\
\hline
Sensor2Text & 73.6 & 68.8 & 63.1 & 53.3 & 74.7 & 38.3 & 33.3 & 62.4 \\
\hline
\end{tabular}
\caption{Captioning scores on the MMAct dataset. Sensor2Text's performance is superior to that of un-finetuned vision LLM models and comparable to that of finetuned models, demonstrating its remarkable ability to use low-information density sensors such as IMU.}
\label{tab:mmact_eval_scores}
\end{table}

\subsection{Privacy}

\begin{figure}
    \centering
    \includegraphics[width=0.7\linewidth]{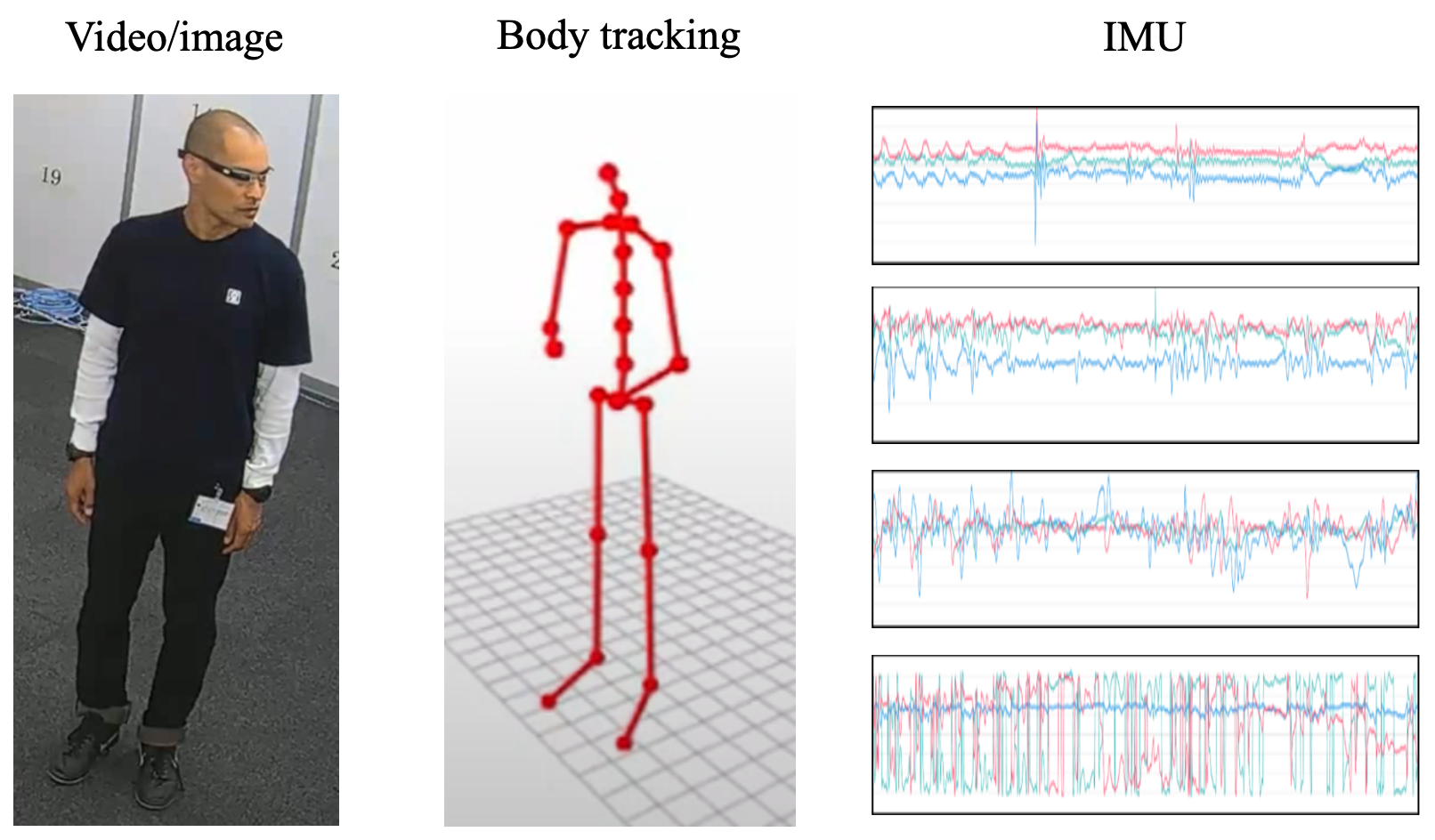}
    \caption{Visualization of information associated with each data modality. Vision captures the most, including subjects' identifiable facial information, clothing, and surroundings. Body tracking captures less information, while IMU sensors capture the least.}
    \label{fig:privacy_diagram}
\end{figure}

One advantage of using sensors to perform daily activity tracking is the preservation of users' privacy. Figure \ref{fig:privacy_diagram} presents a visualization of the decrease in information density between video and various sensor data modalities. As shown, while video captures identifiable information about a subject, wearable sensors are much more privacy-preserving. For example, in one survey \cite{8622110} regarding the privacy of wearable sensors for daily activity tracking and healthcare monitoring, video-based monitoring is identified as more invasive of a subject's surroundings, whereas many privacy concerns regarding wearable sensors can be mitigated with proper data handling and anonymization.

The evaluation using eye-tracking only in ActionSense or privacy-preserving sensors from the MMAct dataset demonstrates that while the model can leverage low-information density sensors such as smartphone and smartwatch IMU to identify users' basic activities, it has reduced performance on detail recognition and recall. Using low information-density sensors would suffice for everyday use, such as inquiring about when a subject sleeps, exercises, or is working on a keyboard. In contrast, for healthcare applications where precision is required, such as inquiring about the type of pill bottle a subject has picked up, high information-density sensors are ideal. Thus, the trade-off between privacy and performance should be carefully considered when deploying such a model for daily activity tracking.

\section{Limitations and Future Work}

Our model demonstrates strong performance when applied to high information-density sensors such as full body-tracking and EMG data than IMU data. However, these can be cumbersome to collect for practical daily use. Additionally, compared to smartphones and smartwatches IMU, full body-tracking data is more invasive of privacy. Thus, the trade-off in performance between employing sensors containing higher information density versus privacy-preserving sensors should be carefully considered when deploying such a model to daily activity tracking. Future work could investigate supplementing smartphone and smartwatch sensors with other privacy-preserving sensors to compensate for the reduced information density while still ensuring strong daily activity-tracking capabilities. Still, we contend that even full-body tracking and EMG sensors are more privacy-preserving than constant video surveillance, as demonstrated in Figure \ref{fig:privacy_diagram}.

Secondly, while the use of foundational vision models and pre-trained large language models greatly enhances the training effectiveness and performance of the Sensor2Text, it also limits the applicability of the model. The use of vision models during training restricts the applicable datasets to those that include vision data for training, which limits the applicability of the model. Furthermore, the use of large language models makes Sensor2Text more computationally expensive compared to classification or captioning models and presents a bottleneck for inference speed. Finally, the applications of the model depend on the licenses of the employed components. In particular, LLaMA 2's community license specifies limits on its commercial use for products with large monthly user counts, and ImageBind forbids commercial use altogether. These can be remedied by selecting an alternative vision model or large language model but nonetheless can present difficulties for the adoption of Sensor2Text. Thus, further work could investigate novel transfer learning methods that potentially reduce the reliance of a sensor-language model on existing vision-language models.

Finally, while Sensor2Text shows the potential to generalize to new users and can be applied in various environments, it struggles with completely unseen environments or activity types. For example, vision-language models like VideoLLaMA and GPT-4 can handle visual footage from new environments without additional training, but Sensor2Text requires further training due to challenges in sensor data collection, such as sensor placements and device variations. The relatively small size of existing sensor-based datasets for daily activity tracking additionally increases the difficulty for a sensor-based model to generalize to new activities. Nonetheless, the strong performance of Sensor2Text with minimal training data suggests it could be a promising alternative to traditional video-based daily activity tracking, addressing privacy and viewpoint concerns. Future research could explore the collection of a large-scale sensor and video dataset for daily activity tracking to train a sensor-language model that can generalize to entirely new activity types.
\section{Conclusion}
In conclusion, this paper introduces a novel, sensor-based approach to enable question-answering on tracking a person's daily activities. Our encoder-decoder model adeptly integrates multiple sensor modalities and text to generate natural language responses, achieving a capability previously only attainable with visual data or a limited number of other modalities to a constrained degree. Despite certain challenges, particularly the scarcity of sensor data and paired sensor/text data, extensive evaluation demonstrates our model's excellent performance across various natural language tasks conditioned on sensor data. This pioneering work highlights the potential of using wearable sensors as an alternative to video in applications such as tracking daily activities, opening up new possibilities for privacy-preserving and unobtrusive monitoring systems.

\bibliographystyle{ACM-Reference-Format}
\bibliography{sample-base}


\begin{thebibliography}{78}


\ifx \showCODEN    \undefined \def \showCODEN     #1{\unskip}     \fi
\ifx \showDOI      \undefined \def \showDOI       #1{#1}\fi
\ifx \showISBNx    \undefined \def \showISBNx     #1{\unskip}     \fi
\ifx \showISBNxiii \undefined \def \showISBNxiii  #1{\unskip}     \fi
\ifx \showISSN     \undefined \def \showISSN      #1{\unskip}     \fi
\ifx \showLCCN     \undefined \def \showLCCN      #1{\unskip}     \fi
\ifx \shownote     \undefined \def \shownote      #1{#1}          \fi
\ifx \showarticletitle \undefined \def \showarticletitle #1{#1}   \fi
\ifx \showURL      \undefined \def \showURL       {\relax}        \fi
\providecommand\bibfield[2]{#2}
\providecommand\bibinfo[2]{#2}
\providecommand\natexlab[1]{#1}
\providecommand\showeprint[2][]{arXiv:#2}

\bibitem[Adib et~al\mbox{.}(2014)]%
        {WiTrack}
\bibfield{author}{\bibinfo{person}{Fadel Adib}, \bibinfo{person}{Zachary Kabelac}, \bibinfo{person}{Dina Katabi}, {and} \bibinfo{person}{Robert~C. Miller}.} \bibinfo{year}{2014}\natexlab{}.
\newblock \showarticletitle{3D tracking via body radio reflections}. In \bibinfo{booktitle}{\emph{Proceedings of the 11th USENIX Conference on Networked Systems Design and Implementation}} (Seattle, WA) \emph{(\bibinfo{series}{NSDI'14})}. \bibinfo{publisher}{USENIX Association}, \bibinfo{address}{USA}, \bibinfo{pages}{317–329}.
\newblock
\showISBNx{9781931971096}


\bibitem[Anderson et~al\mbox{.}(2016)]%
        {anderson2016spice}
\bibfield{author}{\bibinfo{person}{Peter Anderson}, \bibinfo{person}{Basura Fernando}, \bibinfo{person}{Mark Johnson}, {and} \bibinfo{person}{Stephen Gould}.} \bibinfo{year}{2016}\natexlab{}.
\newblock \bibinfo{title}{SPICE: Semantic Propositional Image Caption Evaluation}.
\newblock
\newblock
\showeprint[arxiv]{1607.08822}~[cs.CV]


\bibitem[Anthropic(2024)]%
        {anthropic2024claude}
\bibfield{author}{\bibinfo{person}{Anthropic}.} \bibinfo{year}{2024}\natexlab{}.
\newblock \bibinfo{title}{Claude 3 Family}.
\newblock \bibinfo{howpublished}{\url{https://www.anthropic.com/news/claude-3-family}}.
\newblock
\newblock
\shownote{Accessed: 2024-04-19}.


\bibitem[Antol et~al\mbox{.}(2015)]%
        {VQA}
\bibfield{author}{\bibinfo{person}{Stanislaw Antol}, \bibinfo{person}{Aishwarya Agrawal}, \bibinfo{person}{Jiasen Lu}, \bibinfo{person}{Margaret Mitchell}, \bibinfo{person}{Dhruv Batra}, \bibinfo{person}{C.~Lawrence Zitnick}, {and} \bibinfo{person}{Devi Parikh}.} \bibinfo{year}{2015}\natexlab{}.
\newblock \showarticletitle{{VQA}: {V}isual {Q}uestion {A}nswering}. In \bibinfo{booktitle}{\emph{International Conference on Computer Vision (ICCV)}}.
\newblock


\bibitem[Ba et~al\mbox{.}(2016)]%
        {ba2016layer}
\bibfield{author}{\bibinfo{person}{Jimmy~Lei Ba}, \bibinfo{person}{Jamie~Ryan Kiros}, {and} \bibinfo{person}{Geoffrey~E. Hinton}.} \bibinfo{year}{2016}\natexlab{}.
\newblock \bibinfo{title}{Layer Normalization}.
\newblock
\newblock
\showeprint[arxiv]{1607.06450}~[stat.ML]


\bibitem[Banerjee and Lavie(2005)]%
        {banerjee-lavie-2005-meteor}
\bibfield{author}{\bibinfo{person}{Satanjeev Banerjee} {and} \bibinfo{person}{Alon Lavie}.} \bibinfo{year}{2005}\natexlab{}.
\newblock \showarticletitle{{METEOR}: An Automatic Metric for {MT} Evaluation with Improved Correlation with Human Judgments}. In \bibinfo{booktitle}{\emph{Proceedings of the {ACL} Workshop on Intrinsic and Extrinsic Evaluation Measures for Machine Translation and/or Summarization}}, \bibfield{editor}{\bibinfo{person}{Jade Goldstein}, \bibinfo{person}{Alon Lavie}, \bibinfo{person}{Chin-Yew Lin}, {and} \bibinfo{person}{Clare Voss}} (Eds.). \bibinfo{publisher}{Association for Computational Linguistics}, \bibinfo{address}{Ann Arbor, Michigan}, \bibinfo{pages}{65--72}.
\newblock
\urldef\tempurl%
\url{https://aclanthology.org/W05-0909}
\showURL{%
\tempurl}


\bibitem[Bedri et~al\mbox{.}(2017)]%
        {10.1145/3130902}
\bibfield{author}{\bibinfo{person}{Abdelkareem Bedri}, \bibinfo{person}{Richard Li}, \bibinfo{person}{Malcolm Haynes}, \bibinfo{person}{Raj~Prateek Kosaraju}, \bibinfo{person}{Ishaan Grover}, \bibinfo{person}{Temiloluwa Prioleau}, \bibinfo{person}{Min~Yan Beh}, \bibinfo{person}{Mayank Goel}, \bibinfo{person}{Thad Starner}, {and} \bibinfo{person}{Gregory Abowd}.} \bibinfo{year}{2017}\natexlab{}.
\newblock \showarticletitle{EarBit: Using Wearable Sensors to Detect Eating Episodes in Unconstrained Environments}.
\newblock \bibinfo{journal}{\emph{Proc. ACM Interact. Mob. Wearable Ubiquitous Technol.}} \bibinfo{volume}{1}, \bibinfo{number}{3}, Article \bibinfo{articleno}{37} (\bibinfo{date}{sep} \bibinfo{year}{2017}), \bibinfo{numpages}{20}~pages.
\newblock
\urldef\tempurl%
\url{https://doi.org/10.1145/3130902}
\showDOI{\tempurl}


\bibitem[Brown et~al\mbox{.}(2020)]%
        {brown2020language}
\bibfield{author}{\bibinfo{person}{Tom~B. Brown}, \bibinfo{person}{Benjamin Mann}, \bibinfo{person}{Nick Ryder}, \bibinfo{person}{Melanie Subbiah}, \bibinfo{person}{Jared Kaplan}, \bibinfo{person}{Prafulla Dhariwal}, \bibinfo{person}{Arvind Neelakantan}, \bibinfo{person}{Pranav Shyam}, \bibinfo{person}{Girish Sastry}, \bibinfo{person}{Amanda Askell}, \bibinfo{person}{Sandhini Agarwal}, \bibinfo{person}{Ariel Herbert-Voss}, \bibinfo{person}{Gretchen Krueger}, \bibinfo{person}{Tom Henighan}, \bibinfo{person}{Rewon Child}, \bibinfo{person}{Aditya Ramesh}, \bibinfo{person}{Daniel~M. Ziegler}, \bibinfo{person}{Jeffrey Wu}, \bibinfo{person}{Clemens Winter}, \bibinfo{person}{Christopher Hesse}, \bibinfo{person}{Mark Chen}, \bibinfo{person}{Eric Sigler}, \bibinfo{person}{Mateusz Litwin}, \bibinfo{person}{Scott Gray}, \bibinfo{person}{Benjamin Chess}, \bibinfo{person}{Jack Clark}, \bibinfo{person}{Christopher Berner}, \bibinfo{person}{Sam McCandlish}, \bibinfo{person}{Alec Radford}, \bibinfo{person}{Ilya Sutskever},
  {and} \bibinfo{person}{Dario Amodei}.} \bibinfo{year}{2020}\natexlab{}.
\newblock \bibinfo{title}{Language Models are Few-Shot Learners}.
\newblock
\newblock
\showeprint[arxiv]{2005.14165}~[cs.CL]


\bibitem[Chang et~al\mbox{.}(2020)]%
        {10.1145/3380985}
\bibfield{author}{\bibinfo{person}{Youngjae Chang}, \bibinfo{person}{Akhil Mathur}, \bibinfo{person}{Anton Isopoussu}, \bibinfo{person}{Junehwa Song}, {and} \bibinfo{person}{Fahim Kawsar}.} \bibinfo{year}{2020}\natexlab{}.
\newblock \showarticletitle{A Systematic Study of Unsupervised Domain Adaptation for Robust Human-Activity Recognition}.
\newblock \bibinfo{journal}{\emph{Proc. ACM Interact. Mob. Wearable Ubiquitous Technol.}} \bibinfo{volume}{4}, \bibinfo{number}{1}, Article \bibinfo{articleno}{39} (\bibinfo{date}{mar} \bibinfo{year}{2020}), \bibinfo{numpages}{30}~pages.
\newblock
\urldef\tempurl%
\url{https://doi.org/10.1145/3380985}
\showDOI{\tempurl}


\bibitem[Chen et~al\mbox{.}(2023a)]%
        {chen2023vast}
\bibfield{author}{\bibinfo{person}{Sihan Chen}, \bibinfo{person}{Handong Li}, \bibinfo{person}{Qunbo Wang}, \bibinfo{person}{Zijia Zhao}, \bibinfo{person}{Mingzhen Sun}, \bibinfo{person}{Xinxin Zhu}, {and} \bibinfo{person}{Jing Liu}.} \bibinfo{year}{2023}\natexlab{a}.
\newblock \bibinfo{title}{VAST: A Vision-Audio-Subtitle-Text Omni-Modality Foundation Model and Dataset}.
\newblock
\newblock
\showeprint[arxiv]{2305.18500}~[cs.CV]


\bibitem[Chen et~al\mbox{.}(2021a)]%
        {chen2021viobject}
\bibfield{author}{\bibinfo{person}{Wenqiang Chen}, \bibinfo{person}{Daniel Bevan}, {and} \bibinfo{person}{John Stankovic}.} \bibinfo{year}{2021}\natexlab{a}.
\newblock \showarticletitle{ViObject: A Smartwatch-based Object Recognition System via Vibrations}. In \bibinfo{booktitle}{\emph{Adjunct Proceedings of the 34th Annual ACM Symposium on User Interface Software and Technology}}. \bibinfo{pages}{97--99}.
\newblock


\bibitem[Chen et~al\mbox{.}(2019a)]%
        {chen2019taprint}
\bibfield{author}{\bibinfo{person}{Wenqiang Chen}, \bibinfo{person}{Lin Chen}, \bibinfo{person}{Yandao Huang}, \bibinfo{person}{Xinyu Zhang}, \bibinfo{person}{Lu Wang}, \bibinfo{person}{Rukhsana Ruby}, {and} \bibinfo{person}{Kaishun Wu}.} \bibinfo{year}{2019}\natexlab{a}.
\newblock \showarticletitle{Taprint: Secure text input for commodity smart wristbands}. In \bibinfo{booktitle}{\emph{The 25th Annual International Conference on Mobile Computing and Networking}}. \bibinfo{pages}{1--16}.
\newblock


\bibitem[Chen et~al\mbox{.}(2021b)]%
        {chen2021vifin}
\bibfield{author}{\bibinfo{person}{Wenqiang Chen}, \bibinfo{person}{Lin Chen}, \bibinfo{person}{Meiyi Ma}, \bibinfo{person}{Farshid~Salemi Parizi}, \bibinfo{person}{Shwetak Patel}, {and} \bibinfo{person}{John Stankovic}.} \bibinfo{year}{2021}\natexlab{b}.
\newblock \showarticletitle{ViFin: Harness Passive Vibration to Continuous Micro Finger Writing with a Commodity Smartwatch}.
\newblock \bibinfo{journal}{\emph{Proceedings of the ACM on Interactive, Mobile, Wearable and Ubiquitous Technologies}} \bibinfo{volume}{5}, \bibinfo{number}{1} (\bibinfo{year}{2021}), \bibinfo{pages}{1--25}.
\newblock


\bibitem[Chen et~al\mbox{.}(2020a)]%
        {chen2020continuous}
\bibfield{author}{\bibinfo{person}{Wenqiang Chen}, \bibinfo{person}{Lin Chen}, \bibinfo{person}{Meiyi Ma}, \bibinfo{person}{Farshid~Salemi Parizi}, \bibinfo{person}{Patel Shwetak}, {and} \bibinfo{person}{John Stankovic}.} \bibinfo{year}{2020}\natexlab{a}.
\newblock \showarticletitle{Continuous micro finger writing recognition with a commodity smartwatch: demo abstract}. In \bibinfo{booktitle}{\emph{Proceedings of the 18th Conference on Embedded Networked Sensor Systems}}. \bibinfo{pages}{603--604}.
\newblock


\bibitem[Chen et~al\mbox{.}(2020b)]%
        {chen2020smartwatch}
\bibfield{author}{\bibinfo{person}{Wenqiang Chen}, \bibinfo{person}{Lin Chen}, \bibinfo{person}{Kenneth Wan}, {and} \bibinfo{person}{John Stankovic}.} \bibinfo{year}{2020}\natexlab{b}.
\newblock \showarticletitle{A smartwatch product provides on-body tapping gestures recognition: demo abstract}. In \bibinfo{booktitle}{\emph{Proceedings of the 18th Conference on Embedded Networked Sensor Systems}}. \bibinfo{pages}{589--590}.
\newblock


\bibitem[Chen et~al\mbox{.}(2018)]%
        {chen2018vitype}
\bibfield{author}{\bibinfo{person}{Wenqiang Chen}, \bibinfo{person}{Maoning Guan}, \bibinfo{person}{Yandao Huang}, \bibinfo{person}{Lu Wang}, \bibinfo{person}{Rukhsana Ruby}, \bibinfo{person}{Wen Hu}, {and} \bibinfo{person}{Kaishun Wu}.} \bibinfo{year}{2018}\natexlab{}.
\newblock \showarticletitle{Vitype: A cost efficient on-body typing system through vibration}. In \bibinfo{booktitle}{\emph{2018 15th Annual IEEE International Conference on Sensing, Communication, and Networking (SECON)}}. IEEE, \bibinfo{pages}{1--9}.
\newblock


\bibitem[Chen et~al\mbox{.}(2019b)]%
        {chen2019low}
\bibfield{author}{\bibinfo{person}{Wenqiang Chen}, \bibinfo{person}{Maoning Guan}, \bibinfo{person}{Yandao Huang}, \bibinfo{person}{Lu Wang}, \bibinfo{person}{Rukhsana Ruby}, \bibinfo{person}{Wen Hu}, {and} \bibinfo{person}{Kaishun Wu}.} \bibinfo{year}{2019}\natexlab{b}.
\newblock \showarticletitle{A Low Latency On-Body Typing System through Single Vibration Sensor}.
\newblock \bibinfo{journal}{\emph{IEEE Transactions on Mobile Computing}} \bibinfo{volume}{19}, \bibinfo{number}{11} (\bibinfo{year}{2019}), \bibinfo{pages}{2520--2532}.
\newblock


\bibitem[Chen et~al\mbox{.}(2017a)]%
        {chen2017floc}
\bibfield{author}{\bibinfo{person}{Wenqiang Chen}, \bibinfo{person}{Maoning Guan}, \bibinfo{person}{Lu Wang}, \bibinfo{person}{Rukhsana Ruby}, {and} \bibinfo{person}{Kaishun Wu}.} \bibinfo{year}{2017}\natexlab{a}.
\newblock \showarticletitle{FLoc: Device-free passive indoor localization in complex environments}. In \bibinfo{booktitle}{\emph{2017 IEEE International Conference on Communications (ICC)}}. IEEE, \bibinfo{pages}{1--6}.
\newblock


\bibitem[Chen et~al\mbox{.}(2024a)]%
        {chen2024cavatar}
\bibfield{author}{\bibinfo{person}{Wenqiang Chen}, \bibinfo{person}{Yexin Hu}, \bibinfo{person}{Wei Song}, \bibinfo{person}{Yingcheng Liu}, \bibinfo{person}{Antonio Torralba}, {and} \bibinfo{person}{Wojciech Matusik}.} \bibinfo{year}{2024}\natexlab{a}.
\newblock \showarticletitle{CAvatar: Real-time Human Activity Mesh Reconstruction via Tactile Carpets}.
\newblock \bibinfo{journal}{\emph{Proceedings of the ACM on Interactive, Mobile, Wearable and Ubiquitous Technologies}} \bibinfo{volume}{7}, \bibinfo{number}{4} (\bibinfo{year}{2024}), \bibinfo{pages}{1--24}.
\newblock


\bibitem[Chen et~al\mbox{.}(2017b)]%
        {chen2017virtual}
\bibfield{author}{\bibinfo{person}{Wenqiang Chen}, \bibinfo{person}{Yanming Lian}, \bibinfo{person}{Lu Wang}, \bibinfo{person}{Rukhsana Ruby}, \bibinfo{person}{Wen Hu}, {and} \bibinfo{person}{Kaishun Wu}.} \bibinfo{year}{2017}\natexlab{b}.
\newblock \showarticletitle{Virtual keyboard for wearable wristbands}. In \bibinfo{booktitle}{\emph{Proceedings of the 15th ACM Conference on Embedded Network Sensor Systems}}. \bibinfo{pages}{1--2}.
\newblock


\bibitem[Chen et~al\mbox{.}(2024b)]%
        {10.1145/3643547}
\bibfield{author}{\bibinfo{person}{Wenqiang Chen}, \bibinfo{person}{Shupei Lin}, \bibinfo{person}{Zhencan Peng}, \bibinfo{person}{Farshid~Salemi Parizi}, \bibinfo{person}{Seongkook Heo}, \bibinfo{person}{Shwetak Patel}, \bibinfo{person}{Wojciech Matusik}, \bibinfo{person}{Wei Zhao}, {and} \bibinfo{person}{John Stankovic}.} \bibinfo{year}{2024}\natexlab{b}.
\newblock \showarticletitle{ViObject: Harness Passive Vibrations for Daily Object Recognition with Commodity Smartwatches}.
\newblock \bibinfo{journal}{\emph{Proc. ACM Interact. Mob. Wearable Ubiquitous Technol.}} \bibinfo{volume}{8}, \bibinfo{number}{1}, Article \bibinfo{articleno}{5} (\bibinfo{date}{mar} \bibinfo{year}{2024}), \bibinfo{numpages}{26}~pages.
\newblock
\urldef\tempurl%
\url{https://doi.org/10.1145/3643547}
\showDOI{\tempurl}


\bibitem[CHEN et~al\mbox{.}(2021)]%
        {chen2021sensecollect}
\bibfield{author}{\bibinfo{person}{WENQIANG CHEN}, \bibinfo{person}{SHUPEI LIN}, \bibinfo{person}{ELIZABETH THOMPSON}, {and} \bibinfo{person}{JOHN STANKOVIC}.} \bibinfo{year}{2021}\natexlab{}.
\newblock \showarticletitle{SenseCollect: We Need Efficient Ways to Collect On-body Sensor-based Human Activity Data!}
\newblock \bibinfo{journal}{\emph{Proc. ACM Interact. Mob. Wearable Ubiquitous Technol}} \bibinfo{volume}{1}, \bibinfo{number}{1} (\bibinfo{year}{2021}).
\newblock


\bibitem[Chen and Stankovic(2022)]%
        {chen2022viwatch}
\bibfield{author}{\bibinfo{person}{Wenqiang Chen} {and} \bibinfo{person}{John Stankovic}.} \bibinfo{year}{2022}\natexlab{}.
\newblock \showarticletitle{ViWatch: harness vibrations for finger interactions with commodity smartwatches}. In \bibinfo{booktitle}{\emph{Proceedings of the 13th ACM Wireless of the Students, by the Students, and for the Students Workshop}}. \bibinfo{pages}{4--6}.
\newblock


\bibitem[Chen et~al\mbox{.}(2023b)]%
        {chen2023robust}
\bibfield{author}{\bibinfo{person}{Wenqiang Chen}, \bibinfo{person}{Ziqi Wang}, \bibinfo{person}{Pengrui Quan}, \bibinfo{person}{Zhencan Peng}, \bibinfo{person}{Shupei Lin}, \bibinfo{person}{Mani Srivastava}, \bibinfo{person}{Wojciech Matusik}, {and} \bibinfo{person}{John Stankovic}.} \bibinfo{year}{2023}\natexlab{b}.
\newblock \showarticletitle{Robust Finger Interactions with COTS Smartwatches via Unsupervised Siamese Adaptation}. In \bibinfo{booktitle}{\emph{Proceedings of the 36th Annual ACM Symposium on User Interface Software and Technology}}. \bibinfo{pages}{1--14}.
\newblock


\bibitem[Chen et~al\mbox{.}(2022)]%
        {chen2022making}
\bibfield{author}{\bibinfo{person}{Wenqiang Chen}, \bibinfo{person}{Ziqi Wang}, \bibinfo{person}{Pengrui Quan}, \bibinfo{person}{Zhencan Peng}, \bibinfo{person}{Shupei Lin}, \bibinfo{person}{Mani Srivastava}, {and} \bibinfo{person}{John Stankovic}.} \bibinfo{year}{2022}\natexlab{}.
\newblock \showarticletitle{Making Vibration-based On-body Interaction Robust}. In \bibinfo{booktitle}{\emph{2022 ACM/IEEE 13th International Conference on Cyber-Physical Systems (ICCPS)}}. IEEE, \bibinfo{pages}{300--301}.
\newblock


\bibitem[Chen et~al\mbox{.}(2015)]%
        {chen2015microsoft}
\bibfield{author}{\bibinfo{person}{Xinlei Chen}, \bibinfo{person}{Hao Fang}, \bibinfo{person}{Tsung-Yi Lin}, \bibinfo{person}{Ramakrishna Vedantam}, \bibinfo{person}{Saurabh Gupta}, \bibinfo{person}{Piotr Dollar}, {and} \bibinfo{person}{C.~Lawrence Zitnick}.} \bibinfo{year}{2015}\natexlab{}.
\newblock \bibinfo{title}{Microsoft COCO Captions: Data Collection and Evaluation Server}.
\newblock
\newblock
\showeprint[arxiv]{1504.00325}~[cs.CV]


\bibitem[Datta et~al\mbox{.}(2018)]%
        {8622110}
\bibfield{author}{\bibinfo{person}{Prerit Datta}, \bibinfo{person}{Akbar~Siami Namin}, {and} \bibinfo{person}{Moitrayee Chatterjee}.} \bibinfo{year}{2018}\natexlab{}.
\newblock \showarticletitle{A Survey of Privacy Concerns in Wearable Devices}. In \bibinfo{booktitle}{\emph{2018 IEEE International Conference on Big Data (Big Data)}}. \bibinfo{pages}{4549--4553}.
\newblock
\urldef\tempurl%
\url{https://doi.org/10.1109/BigData.2018.8622110}
\showDOI{\tempurl}


\bibitem[Deldari et~al\mbox{.}(2022)]%
        {10.1145/3550316}
\bibfield{author}{\bibinfo{person}{Shohreh Deldari}, \bibinfo{person}{Hao Xue}, \bibinfo{person}{Aaqib Saeed}, \bibinfo{person}{Daniel~V. Smith}, {and} \bibinfo{person}{Flora~D. Salim}.} \bibinfo{year}{2022}\natexlab{}.
\newblock \showarticletitle{COCOA: Cross Modality Contrastive Learning for Sensor Data}.
\newblock \bibinfo{journal}{\emph{Proc. ACM Interact. Mob. Wearable Ubiquitous Technol.}} \bibinfo{volume}{6}, \bibinfo{number}{3}, Article \bibinfo{articleno}{108} (\bibinfo{date}{sep} \bibinfo{year}{2022}), \bibinfo{numpages}{28}~pages.
\newblock
\urldef\tempurl%
\url{https://doi.org/10.1145/3550316}
\showDOI{\tempurl}


\bibitem[DelPreto et~al\mbox{.}(2022)]%
        {delpretoLiu2022actionSense}
\bibfield{author}{\bibinfo{person}{Joseph DelPreto}, \bibinfo{person}{Chao Liu}, \bibinfo{person}{Yiyue Luo}, \bibinfo{person}{Michael Foshey}, \bibinfo{person}{Yunzhu Li}, \bibinfo{person}{Antonio Torralba}, \bibinfo{person}{Wojciech Matusik}, {and} \bibinfo{person}{Daniela Rus}.} \bibinfo{year}{2022}\natexlab{}.
\newblock \showarticletitle{{ActionSense}: A Multimodal Dataset and Recording Framework for Human Activities Using Wearable Sensors in a Kitchen Environment}. In \bibinfo{booktitle}{\emph{Neural Information Processing Systems (NeurIPS) Track on Datasets and Benchmarks}}.
\newblock
\urldef\tempurl%
\url{https://action-sense.csail.mit.edu}
\showURL{%
\tempurl}


\bibitem[Fan et~al\mbox{.}(2020)]%
        {fan2020inhome}
\bibfield{author}{\bibinfo{person}{Lijie Fan}, \bibinfo{person}{Tianhong Li}, \bibinfo{person}{Yuan Yuan}, {and} \bibinfo{person}{Dina Katabi}.} \bibinfo{year}{2020}\natexlab{}.
\newblock \bibinfo{title}{In-Home Daily-Life Captioning Using Radio Signals}.
\newblock
\newblock
\showeprint[arxiv]{2008.10966}~[cs.CV]


\bibitem[Fligge et~al\mbox{.}(2013)]%
        {FLIGGE2013402}
\bibfield{author}{\bibinfo{person}{Nadine Fligge}, \bibinfo{person}{Holger Urbanek}, {and} \bibinfo{person}{Patrick {van der Smagt}}.} \bibinfo{year}{2013}\natexlab{}.
\newblock \showarticletitle{Relation between object properties and EMG during reaching to grasp}.
\newblock \bibinfo{journal}{\emph{Journal of Electromyography and Kinesiology}} \bibinfo{volume}{23}, \bibinfo{number}{2} (\bibinfo{year}{2013}), \bibinfo{pages}{402--410}.
\newblock
\showISSN{1050-6411}
\urldef\tempurl%
\url{https://doi.org/10.1016/j.jelekin.2012.10.010}
\showDOI{\tempurl}


\bibitem[Ghaderi et~al\mbox{.}(2022)]%
        {ghaderi2022diverse}
\bibfield{author}{\bibinfo{person}{Zohreh Ghaderi}, \bibinfo{person}{Leonard Salewski}, {and} \bibinfo{person}{Hendrik P.~A. Lensch}.} \bibinfo{year}{2022}\natexlab{}.
\newblock \bibinfo{title}{Diverse Video Captioning by Adaptive Spatio-temporal Attention}.
\newblock
\newblock
\showeprint[arxiv]{2208.09266}~[cs.CV]


\bibitem[Girdhar et~al\mbox{.}(2023)]%
        {girdhar2023imagebind}
\bibfield{author}{\bibinfo{person}{Rohit Girdhar}, \bibinfo{person}{Alaaeldin El-Nouby}, \bibinfo{person}{Zhuang Liu}, \bibinfo{person}{Mannat Singh}, \bibinfo{person}{Kalyan~Vasudev Alwala}, \bibinfo{person}{Armand Joulin}, {and} \bibinfo{person}{Ishan Misra}.} \bibinfo{year}{2023}\natexlab{}.
\newblock \bibinfo{title}{ImageBind: One Embedding Space To Bind Them All}.
\newblock
\newblock
\showeprint[arxiv]{2305.05665}~[cs.CV]


\bibitem[Guan et~al\mbox{.}(2019)]%
        {guan2019faceinput}
\bibfield{author}{\bibinfo{person}{Maoning Guan}, \bibinfo{person}{Wenqiang Chen}, \bibinfo{person}{Yandao Huang}, \bibinfo{person}{Rukhsana Ruby}, {and} \bibinfo{person}{Kaishun Wu}.} \bibinfo{year}{2019}\natexlab{}.
\newblock \showarticletitle{FaceInput: a hand-free and secure text entry system through facial vibration}. In \bibinfo{booktitle}{\emph{2019 16th Annual IEEE International Conference on Sensing, Communication, and Networking (SECON)}}. IEEE, \bibinfo{pages}{1--9}.
\newblock


\bibitem[Guan and Pl\"{o}tz(2017)]%
        {10.1145/3090076}
\bibfield{author}{\bibinfo{person}{Yu Guan} {and} \bibinfo{person}{Thomas Pl\"{o}tz}.} \bibinfo{year}{2017}\natexlab{}.
\newblock \showarticletitle{Ensembles of Deep LSTM Learners for Activity Recognition using Wearables}.
\newblock \bibinfo{journal}{\emph{Proc. ACM Interact. Mob. Wearable Ubiquitous Technol.}} \bibinfo{volume}{1}, \bibinfo{number}{2}, Article \bibinfo{articleno}{11} (\bibinfo{date}{jun} \bibinfo{year}{2017}), \bibinfo{numpages}{28}~pages.
\newblock
\urldef\tempurl%
\url{https://doi.org/10.1145/3090076}
\showDOI{\tempurl}


\bibitem[Guo et~al\mbox{.}(2023)]%
        {guo2023pointbind}
\bibfield{author}{\bibinfo{person}{Ziyu Guo}, \bibinfo{person}{Renrui Zhang}, \bibinfo{person}{Xiangyang Zhu}, \bibinfo{person}{Yiwen Tang}, \bibinfo{person}{Xianzheng Ma}, \bibinfo{person}{Jiaming Han}, \bibinfo{person}{Kexin Chen}, \bibinfo{person}{Peng Gao}, \bibinfo{person}{Xianzhi Li}, \bibinfo{person}{Hongsheng Li}, {and} \bibinfo{person}{Pheng-Ann Heng}.} \bibinfo{year}{2023}\natexlab{}.
\newblock \bibinfo{title}{Point-Bind \& Point-LLM: Aligning Point Cloud with Multi-modality for 3D Understanding, Generation, and Instruction Following}.
\newblock
\newblock
\showeprint[arxiv]{2309.00615}~[cs.CV]


\bibitem[Han et~al\mbox{.}(2023)]%
        {han2023imagebindllm}
\bibfield{author}{\bibinfo{person}{Jiaming Han}, \bibinfo{person}{Renrui Zhang}, \bibinfo{person}{Wenqi Shao}, \bibinfo{person}{Peng Gao}, \bibinfo{person}{Peng Xu}, \bibinfo{person}{Han Xiao}, \bibinfo{person}{Kaipeng Zhang}, \bibinfo{person}{Chris Liu}, \bibinfo{person}{Song Wen}, \bibinfo{person}{Ziyu Guo}, \bibinfo{person}{Xudong Lu}, \bibinfo{person}{Shuai Ren}, \bibinfo{person}{Yafei Wen}, \bibinfo{person}{Xiaoxin Chen}, \bibinfo{person}{Xiangyu Yue}, \bibinfo{person}{Hongsheng Li}, {and} \bibinfo{person}{Yu Qiao}.} \bibinfo{year}{2023}\natexlab{}.
\newblock \bibinfo{title}{ImageBind-LLM: Multi-modality Instruction Tuning}.
\newblock
\newblock
\showeprint[arxiv]{2309.03905}~[cs.MM]


\bibitem[Haresamudram et~al\mbox{.}(2021)]%
        {10.1145/3463506}
\bibfield{author}{\bibinfo{person}{Harish Haresamudram}, \bibinfo{person}{Irfan Essa}, {and} \bibinfo{person}{Thomas Pl\"{o}tz}.} \bibinfo{year}{2021}\natexlab{}.
\newblock \showarticletitle{Contrastive Predictive Coding for Human Activity Recognition}.
\newblock \bibinfo{journal}{\emph{Proc. ACM Interact. Mob. Wearable Ubiquitous Technol.}} \bibinfo{volume}{5}, \bibinfo{number}{2}, Article \bibinfo{articleno}{65} (\bibinfo{date}{jun} \bibinfo{year}{2021}), \bibinfo{numpages}{26}~pages.
\newblock
\urldef\tempurl%
\url{https://doi.org/10.1145/3463506}
\showDOI{\tempurl}


\bibitem[Huang et~al\mbox{.}(2019)]%
        {huang2019g}
\bibfield{author}{\bibinfo{person}{Yandao Huang}, \bibinfo{person}{Wenqiang Chen}, \bibinfo{person}{Hongjie Chen}, \bibinfo{person}{Lu Wang}, {and} \bibinfo{person}{Kaishun Wu}.} \bibinfo{year}{2019}\natexlab{}.
\newblock \showarticletitle{G-fall: device-free and training-free fall detection with geophones}. In \bibinfo{booktitle}{\emph{2019 16th Annual IEEE International Conference on Sensing, Communication, and Networking (SECON)}}. IEEE, \bibinfo{pages}{1--9}.
\newblock


\bibitem[Kim et~al\mbox{.}(2024)]%
        {kim2024healthllmlargelanguagemodels}
\bibfield{author}{\bibinfo{person}{Yubin Kim}, \bibinfo{person}{Xuhai Xu}, \bibinfo{person}{Daniel McDuff}, \bibinfo{person}{Cynthia Breazeal}, {and} \bibinfo{person}{Hae~Won Park}.} \bibinfo{year}{2024}\natexlab{}.
\newblock \bibinfo{title}{Health-LLM: Large Language Models for Health Prediction via Wearable Sensor Data}.
\newblock
\newblock
\showeprint[arxiv]{2401.06866}~[cs.CL]
\urldef\tempurl%
\url{https://arxiv.org/abs/2401.06866}
\showURL{%
\tempurl}


\bibitem[King et~al\mbox{.}(2024)]%
        {10.1145/3643505}
\bibfield{author}{\bibinfo{person}{Evan King}, \bibinfo{person}{Haoxiang Yu}, \bibinfo{person}{Sangsu Lee}, {and} \bibinfo{person}{Christine Julien}.} \bibinfo{year}{2024}\natexlab{}.
\newblock \showarticletitle{Sasha: Creative Goal-Oriented Reasoning in Smart Homes with Large Language Models}.
\newblock \bibinfo{journal}{\emph{Proc. ACM Interact. Mob. Wearable Ubiquitous Technol.}} \bibinfo{volume}{8}, \bibinfo{number}{1}, Article \bibinfo{articleno}{12} (\bibinfo{date}{mar} \bibinfo{year}{2024}), \bibinfo{numpages}{38}~pages.
\newblock
\urldef\tempurl%
\url{https://doi.org/10.1145/3643505}
\showDOI{\tempurl}


\bibitem[Kingma and Ba(2017)]%
        {kingma2017adam}
\bibfield{author}{\bibinfo{person}{Diederik~P. Kingma} {and} \bibinfo{person}{Jimmy Ba}.} \bibinfo{year}{2017}\natexlab{}.
\newblock \bibinfo{title}{Adam: A Method for Stochastic Optimization}.
\newblock
\newblock
\showeprint[arxiv]{1412.6980}~[cs.LG]


\bibitem[Kong et~al\mbox{.}(2019a)]%
        {kong2019mmact}
\bibfield{author}{\bibinfo{person}{Quan Kong}, \bibinfo{person}{Ziming Wu}, \bibinfo{person}{Ziwei Deng}, \bibinfo{person}{Martin Klinkigt}, \bibinfo{person}{Bin Tong}, {and} \bibinfo{person}{Tomokazu Murakami}.} \bibinfo{year}{2019}\natexlab{a}.
\newblock \showarticletitle{MMAct: A large-scale dataset for cross modal human action understanding}. In \bibinfo{booktitle}{\emph{2019 IEEE/CVF International Conference on Computer Vision (ICCV)}}. IEEE, \bibinfo{pages}{8657--8666}.
\newblock
\urldef\tempurl%
\url{https://doi.org/10.1109/ICCV.2019.00875}
\showDOI{\tempurl}


\bibitem[Kong et~al\mbox{.}(2019b)]%
        {9009579}
\bibfield{author}{\bibinfo{person}{Quan Kong}, \bibinfo{person}{Ziming Wu}, \bibinfo{person}{Ziwei Deng}, \bibinfo{person}{Martin Klinkigt}, \bibinfo{person}{Bin Tong}, {and} \bibinfo{person}{Tomokazu Murakami}.} \bibinfo{year}{2019}\natexlab{b}.
\newblock \showarticletitle{MMAct: A Large-Scale Dataset for Cross Modal Human Action Understanding}. In \bibinfo{booktitle}{\emph{2019 IEEE/CVF International Conference on Computer Vision (ICCV)}}. \bibinfo{pages}{8657--8666}.
\newblock
\urldef\tempurl%
\url{https://doi.org/10.1109/ICCV.2019.00875}
\showDOI{\tempurl}


\bibitem[Labs(2022)]%
        {PupilCore}
\bibfield{author}{\bibinfo{person}{Pupil Labs}.} \bibinfo{year}{2022}\natexlab{}.
\newblock \bibinfo{title}{Pupil Core eye-tracking headset}.
\newblock \bibinfo{howpublished}{\url{https://pupil-labs.com/products/core}}.
\newblock


\bibitem[Li et~al\mbox{.}(2023)]%
        {li2023blip2}
\bibfield{author}{\bibinfo{person}{Junnan Li}, \bibinfo{person}{Dongxu Li}, \bibinfo{person}{Silvio Savarese}, {and} \bibinfo{person}{Steven Hoi}.} \bibinfo{year}{2023}\natexlab{}.
\newblock \bibinfo{title}{BLIP-2: Bootstrapping Language-Image Pre-training with Frozen Image Encoders and Large Language Models}.
\newblock
\newblock
\showeprint[arxiv]{2301.12597}~[cs.CV]


\bibitem[Li et~al\mbox{.}(2019)]%
        {li2019making}
\bibfield{author}{\bibinfo{person}{Tianhong Li}, \bibinfo{person}{Lijie Fan}, \bibinfo{person}{Mingmin Zhao}, \bibinfo{person}{Yingcheng Liu}, {and} \bibinfo{person}{Dina Katabi}.} \bibinfo{year}{2019}\natexlab{}.
\newblock \bibinfo{title}{Making the Invisible Visible: Action Recognition Through Walls and Occlusions}.
\newblock
\newblock
\showeprint[arxiv]{1909.09300}~[cs.CV]


\bibitem[Li et~al\mbox{.}(2017)]%
        {10.1145/3130940}
\bibfield{author}{\bibinfo{person}{Xiang Li}, \bibinfo{person}{Daqing Zhang}, \bibinfo{person}{Qin Lv}, \bibinfo{person}{Jie Xiong}, \bibinfo{person}{Shengjie Li}, \bibinfo{person}{Yue Zhang}, {and} \bibinfo{person}{Hong Mei}.} \bibinfo{year}{2017}\natexlab{}.
\newblock \showarticletitle{IndoTrack: Device-Free Indoor Human Tracking with Commodity Wi-Fi}.
\newblock \bibinfo{journal}{\emph{Proc. ACM Interact. Mob. Wearable Ubiquitous Technol.}} \bibinfo{volume}{1}, \bibinfo{number}{3}, Article \bibinfo{articleno}{72} (\bibinfo{date}{sep} \bibinfo{year}{2017}), \bibinfo{numpages}{22}~pages.
\newblock
\urldef\tempurl%
\url{https://doi.org/10.1145/3130940}
\showDOI{\tempurl}


\bibitem[Liang et~al\mbox{.}(2024)]%
        {10272675}
\bibfield{author}{\bibinfo{person}{Yuanzhi Liang}, \bibinfo{person}{Linchao Zhu}, \bibinfo{person}{Xiaohan Wang}, {and} \bibinfo{person}{Yi Yang}.} \bibinfo{year}{2024}\natexlab{}.
\newblock \showarticletitle{IcoCap: Improving Video Captioning by Compounding Images}.
\newblock \bibinfo{journal}{\emph{IEEE Transactions on Multimedia}}  \bibinfo{volume}{26} (\bibinfo{year}{2024}), \bibinfo{pages}{4389--4400}.
\newblock
\urldef\tempurl%
\url{https://doi.org/10.1109/TMM.2023.3322329}
\showDOI{\tempurl}


\bibitem[Lin(2004)]%
        {lin-2004-rouge}
\bibfield{author}{\bibinfo{person}{Chin-Yew Lin}.} \bibinfo{year}{2004}\natexlab{}.
\newblock \showarticletitle{{ROUGE}: A Package for Automatic Evaluation of Summaries}. In \bibinfo{booktitle}{\emph{Text Summarization Branches Out}}. \bibinfo{publisher}{Association for Computational Linguistics}, \bibinfo{address}{Barcelona, Spain}, \bibinfo{pages}{74--81}.
\newblock
\urldef\tempurl%
\url{https://aclanthology.org/W04-1013}
\showURL{%
\tempurl}


\bibitem[Liu et~al\mbox{.}(2023)]%
        {liu2023visual}
\bibfield{author}{\bibinfo{person}{Haotian Liu}, \bibinfo{person}{Chunyuan Li}, \bibinfo{person}{Qingyang Wu}, {and} \bibinfo{person}{Yong~Jae Lee}.} \bibinfo{year}{2023}\natexlab{}.
\newblock \bibinfo{title}{Visual Instruction Tuning}.
\newblock
\newblock
\showeprint[arxiv]{2304.08485}~[cs.CV]


\bibitem[Lu et~al\mbox{.}(2022)]%
        {lu2022learn}
\bibfield{author}{\bibinfo{person}{Pan Lu}, \bibinfo{person}{Swaroop Mishra}, \bibinfo{person}{Tony Xia}, \bibinfo{person}{Liang Qiu}, \bibinfo{person}{Kai-Wei Chang}, \bibinfo{person}{Song-Chun Zhu}, \bibinfo{person}{Oyvind Tafjord}, \bibinfo{person}{Peter Clark}, {and} \bibinfo{person}{Ashwin Kalyan}.} \bibinfo{year}{2022}\natexlab{}.
\newblock \showarticletitle{Learn to Explain: Multimodal Reasoning via Thought Chains for Science Question Answering}. In \bibinfo{booktitle}{\emph{The 36th Conference on Neural Information Processing Systems (NeurIPS)}}.
\newblock


\bibitem[Luo et~al\mbox{.}(2020)]%
        {luo2020univl}
\bibfield{author}{\bibinfo{person}{Huaishao Luo}, \bibinfo{person}{Lei Ji}, \bibinfo{person}{Botian Shi}, \bibinfo{person}{Haoyang Huang}, \bibinfo{person}{Nan Duan}, \bibinfo{person}{Tianrui Li}, \bibinfo{person}{Jason Li}, \bibinfo{person}{Taroon Bharti}, {and} \bibinfo{person}{Ming Zhou}.} \bibinfo{year}{2020}\natexlab{}.
\newblock \bibinfo{title}{UniVL: A Unified Video and Language Pre-Training Model for Multimodal Understanding and Generation}.
\newblock
\newblock
\showeprint[arxiv]{2002.06353}~[cs.CV]


\bibitem[Ma et~al\mbox{.}(2021)]%
        {10.1145/3448074}
\bibfield{author}{\bibinfo{person}{Haojie Ma}, \bibinfo{person}{Zhijie Zhang}, \bibinfo{person}{Wenzhong Li}, {and} \bibinfo{person}{Sanglu Lu}.} \bibinfo{year}{2021}\natexlab{}.
\newblock \showarticletitle{Unsupervised Human Activity Representation Learning with Multi-task Deep Clustering}.
\newblock \bibinfo{journal}{\emph{Proc. ACM Interact. Mob. Wearable Ubiquitous Technol.}} \bibinfo{volume}{5}, \bibinfo{number}{1}, Article \bibinfo{articleno}{48} (\bibinfo{date}{mar} \bibinfo{year}{2021}), \bibinfo{numpages}{25}~pages.
\newblock
\urldef\tempurl%
\url{https://doi.org/10.1145/3448074}
\showDOI{\tempurl}


\bibitem[Miao et~al\mbox{.}(2022)]%
        {10.1145/3550331}
\bibfield{author}{\bibinfo{person}{Shenghuan Miao}, \bibinfo{person}{Ling Chen}, \bibinfo{person}{Rong Hu}, {and} \bibinfo{person}{Yingsong Luo}.} \bibinfo{year}{2022}\natexlab{}.
\newblock \showarticletitle{Towards a Dynamic Inter-Sensor Correlations Learning Framework for Multi-Sensor-Based Wearable Human Activity Recognition}.
\newblock \bibinfo{journal}{\emph{Proc. ACM Interact. Mob. Wearable Ubiquitous Technol.}} \bibinfo{volume}{6}, \bibinfo{number}{3}, Article \bibinfo{articleno}{130} (\bibinfo{date}{sep} \bibinfo{year}{2022}), \bibinfo{numpages}{25}~pages.
\newblock
\urldef\tempurl%
\url{https://doi.org/10.1145/3550331}
\showDOI{\tempurl}


\bibitem[Ni et~al\mbox{.}(2021)]%
        {ni2021large}
\bibfield{author}{\bibinfo{person}{Jianmo Ni}, \bibinfo{person}{Chen Qu}, \bibinfo{person}{Jing Lu}, \bibinfo{person}{Zhuyun Dai}, \bibinfo{person}{Gustavo~Hernández Ábrego}, \bibinfo{person}{Ji Ma}, \bibinfo{person}{Vincent~Y. Zhao}, \bibinfo{person}{Yi Luan}, \bibinfo{person}{Keith~B. Hall}, \bibinfo{person}{Ming-Wei Chang}, {and} \bibinfo{person}{Yinfei Yang}.} \bibinfo{year}{2021}\natexlab{}.
\newblock \bibinfo{title}{Large Dual Encoders Are Generalizable Retrievers}.
\newblock
\newblock
\showeprint[arxiv]{2112.07899}~[cs.IR]


\bibitem[Niu et~al\mbox{.}(2024)]%
        {niu2024screenagent}
\bibfield{author}{\bibinfo{person}{Runliang Niu}, \bibinfo{person}{Jindong Li}, \bibinfo{person}{Shiqi Wang}, \bibinfo{person}{Yali Fu}, \bibinfo{person}{Xiyu Hu}, \bibinfo{person}{Xueyuan Leng}, \bibinfo{person}{He Kong}, \bibinfo{person}{Yi Chang}, {and} \bibinfo{person}{Qi Wang}.} \bibinfo{year}{2024}\natexlab{}.
\newblock \bibinfo{title}{ScreenAgent: A Vision Language Model-driven Computer Control Agent}.
\newblock
\newblock
\showeprint[arxiv]{2402.07945}~[cs.HC]


\bibitem[OpenAI et~al\mbox{.}(2024)]%
        {openai2024gpt4}
\bibfield{author}{\bibinfo{person}{OpenAI}, \bibinfo{person}{Josh Achiam}, \bibinfo{person}{Steven Adler}, \bibinfo{person}{Sandhini Agarwal}, \bibinfo{person}{Lama Ahmad}, \bibinfo{person}{Ilge Akkaya}, \bibinfo{person}{Florencia~Leoni Aleman}, \bibinfo{person}{Diogo Almeida}, \bibinfo{person}{Janko Altenschmidt}, \bibinfo{person}{Sam Altman}, \bibinfo{person}{Shyamal Anadkat}, \bibinfo{person}{Red Avila}, \bibinfo{person}{Igor Babuschkin}, \bibinfo{person}{Suchir Balaji}, \bibinfo{person}{Valerie Balcom}, \bibinfo{person}{Paul Baltescu}, \bibinfo{person}{Haiming Bao}, \bibinfo{person}{Mohammad Bavarian}, \bibinfo{person}{Jeff Belgum}, \bibinfo{person}{Irwan Bello}, \bibinfo{person}{Jake Berdine}, \bibinfo{person}{Gabriel Bernadett-Shapiro}, \bibinfo{person}{Christopher Berner}, \bibinfo{person}{Lenny Bogdonoff}, \bibinfo{person}{Oleg Boiko}, \bibinfo{person}{Madelaine Boyd}, \bibinfo{person}{Anna-Luisa Brakman}, \bibinfo{person}{Greg Brockman}, \bibinfo{person}{Tim Brooks}, \bibinfo{person}{Miles Brundage},
  \bibinfo{person}{Kevin Button}, \bibinfo{person}{Trevor Cai}, \bibinfo{person}{Rosie Campbell}, \bibinfo{person}{Andrew Cann}, \bibinfo{person}{Brittany Carey}, \bibinfo{person}{Chelsea Carlson}, \bibinfo{person}{Rory Carmichael}, \bibinfo{person}{Brooke Chan}, \bibinfo{person}{Che Chang}, \bibinfo{person}{Fotis Chantzis}, \bibinfo{person}{Derek Chen}, \bibinfo{person}{Sully Chen}, \bibinfo{person}{Ruby Chen}, \bibinfo{person}{Jason Chen}, \bibinfo{person}{Mark Chen}, \bibinfo{person}{Ben Chess}, \bibinfo{person}{Chester Cho}, \bibinfo{person}{Casey Chu}, \bibinfo{person}{Hyung~Won Chung}, \bibinfo{person}{Dave Cummings}, \bibinfo{person}{Jeremiah Currier}, \bibinfo{person}{Yunxing Dai}, \bibinfo{person}{Cory Decareaux}, \bibinfo{person}{Thomas Degry}, \bibinfo{person}{Noah Deutsch}, \bibinfo{person}{Damien Deville}, \bibinfo{person}{Arka Dhar}, \bibinfo{person}{David Dohan}, \bibinfo{person}{Steve Dowling}, \bibinfo{person}{Sheila Dunning}, \bibinfo{person}{Adrien Ecoffet}, \bibinfo{person}{Atty Eleti},
  \bibinfo{person}{Tyna Eloundou}, \bibinfo{person}{David Farhi}, \bibinfo{person}{Liam Fedus}, \bibinfo{person}{Niko Felix}, \bibinfo{person}{Simón~Posada Fishman}, \bibinfo{person}{Juston Forte}, \bibinfo{person}{Isabella Fulford}, \bibinfo{person}{Leo Gao}, \bibinfo{person}{Elie Georges}, \bibinfo{person}{Christian Gibson}, \bibinfo{person}{Vik Goel}, \bibinfo{person}{Tarun Gogineni}, \bibinfo{person}{Gabriel Goh}, \bibinfo{person}{Rapha Gontijo-Lopes}, \bibinfo{person}{Jonathan Gordon}, \bibinfo{person}{Morgan Grafstein}, \bibinfo{person}{Scott Gray}, \bibinfo{person}{Ryan Greene}, \bibinfo{person}{Joshua Gross}, \bibinfo{person}{Shixiang~Shane Gu}, \bibinfo{person}{Yufei Guo}, \bibinfo{person}{Chris Hallacy}, \bibinfo{person}{Jesse Han}, \bibinfo{person}{Jeff Harris}, \bibinfo{person}{Yuchen He}, \bibinfo{person}{Mike Heaton}, \bibinfo{person}{Johannes Heidecke}, \bibinfo{person}{Chris Hesse}, \bibinfo{person}{Alan Hickey}, \bibinfo{person}{Wade Hickey}, \bibinfo{person}{Peter Hoeschele},
  \bibinfo{person}{Brandon Houghton}, \bibinfo{person}{Kenny Hsu}, \bibinfo{person}{Shengli Hu}, \bibinfo{person}{Xin Hu}, \bibinfo{person}{Joost Huizinga}, \bibinfo{person}{Shantanu Jain}, \bibinfo{person}{Shawn Jain}, \bibinfo{person}{Joanne Jang}, \bibinfo{person}{Angela Jiang}, \bibinfo{person}{Roger Jiang}, \bibinfo{person}{Haozhun Jin}, \bibinfo{person}{Denny Jin}, \bibinfo{person}{Shino Jomoto}, \bibinfo{person}{Billie Jonn}, \bibinfo{person}{Heewoo Jun}, \bibinfo{person}{Tomer Kaftan}, \bibinfo{person}{Łukasz Kaiser}, \bibinfo{person}{Ali Kamali}, \bibinfo{person}{Ingmar Kanitscheider}, \bibinfo{person}{Nitish~Shirish Keskar}, \bibinfo{person}{Tabarak Khan}, \bibinfo{person}{Logan Kilpatrick}, \bibinfo{person}{Jong~Wook Kim}, \bibinfo{person}{Christina Kim}, \bibinfo{person}{Yongjik Kim}, \bibinfo{person}{Jan~Hendrik Kirchner}, \bibinfo{person}{Jamie Kiros}, \bibinfo{person}{Matt Knight}, \bibinfo{person}{Daniel Kokotajlo}, \bibinfo{person}{Łukasz Kondraciuk}, \bibinfo{person}{Andrew Kondrich},
  \bibinfo{person}{Aris Konstantinidis}, \bibinfo{person}{Kyle Kosic}, \bibinfo{person}{Gretchen Krueger}, \bibinfo{person}{Vishal Kuo}, \bibinfo{person}{Michael Lampe}, \bibinfo{person}{Ikai Lan}, \bibinfo{person}{Teddy Lee}, \bibinfo{person}{Jan Leike}, \bibinfo{person}{Jade Leung}, \bibinfo{person}{Daniel Levy}, \bibinfo{person}{Chak~Ming Li}, \bibinfo{person}{Rachel Lim}, \bibinfo{person}{Molly Lin}, \bibinfo{person}{Stephanie Lin}, \bibinfo{person}{Mateusz Litwin}, \bibinfo{person}{Theresa Lopez}, \bibinfo{person}{Ryan Lowe}, \bibinfo{person}{Patricia Lue}, \bibinfo{person}{Anna Makanju}, \bibinfo{person}{Kim Malfacini}, \bibinfo{person}{Sam Manning}, \bibinfo{person}{Todor Markov}, \bibinfo{person}{Yaniv Markovski}, \bibinfo{person}{Bianca Martin}, \bibinfo{person}{Katie Mayer}, \bibinfo{person}{Andrew Mayne}, \bibinfo{person}{Bob McGrew}, \bibinfo{person}{Scott~Mayer McKinney}, \bibinfo{person}{Christine McLeavey}, \bibinfo{person}{Paul McMillan}, \bibinfo{person}{Jake McNeil}, \bibinfo{person}{David
  Medina}, \bibinfo{person}{Aalok Mehta}, \bibinfo{person}{Jacob Menick}, \bibinfo{person}{Luke Metz}, \bibinfo{person}{Andrey Mishchenko}, \bibinfo{person}{Pamela Mishkin}, \bibinfo{person}{Vinnie Monaco}, \bibinfo{person}{Evan Morikawa}, \bibinfo{person}{Daniel Mossing}, \bibinfo{person}{Tong Mu}, \bibinfo{person}{Mira Murati}, \bibinfo{person}{Oleg Murk}, \bibinfo{person}{David Mély}, \bibinfo{person}{Ashvin Nair}, \bibinfo{person}{Reiichiro Nakano}, \bibinfo{person}{Rajeev Nayak}, \bibinfo{person}{Arvind Neelakantan}, \bibinfo{person}{Richard Ngo}, \bibinfo{person}{Hyeonwoo Noh}, \bibinfo{person}{Long Ouyang}, \bibinfo{person}{Cullen O'Keefe}, \bibinfo{person}{Jakub Pachocki}, \bibinfo{person}{Alex Paino}, \bibinfo{person}{Joe Palermo}, \bibinfo{person}{Ashley Pantuliano}, \bibinfo{person}{Giambattista Parascandolo}, \bibinfo{person}{Joel Parish}, \bibinfo{person}{Emy Parparita}, \bibinfo{person}{Alex Passos}, \bibinfo{person}{Mikhail Pavlov}, \bibinfo{person}{Andrew Peng}, \bibinfo{person}{Adam
  Perelman}, \bibinfo{person}{Filipe de Avila Belbute~Peres}, \bibinfo{person}{Michael Petrov}, \bibinfo{person}{Henrique~Ponde de Oliveira~Pinto}, \bibinfo{person}{Michael}, \bibinfo{person}{Pokorny}, \bibinfo{person}{Michelle Pokrass}, \bibinfo{person}{Vitchyr~H. Pong}, \bibinfo{person}{Tolly Powell}, \bibinfo{person}{Alethea Power}, \bibinfo{person}{Boris Power}, \bibinfo{person}{Elizabeth Proehl}, \bibinfo{person}{Raul Puri}, \bibinfo{person}{Alec Radford}, \bibinfo{person}{Jack Rae}, \bibinfo{person}{Aditya Ramesh}, \bibinfo{person}{Cameron Raymond}, \bibinfo{person}{Francis Real}, \bibinfo{person}{Kendra Rimbach}, \bibinfo{person}{Carl Ross}, \bibinfo{person}{Bob Rotsted}, \bibinfo{person}{Henri Roussez}, \bibinfo{person}{Nick Ryder}, \bibinfo{person}{Mario Saltarelli}, \bibinfo{person}{Ted Sanders}, \bibinfo{person}{Shibani Santurkar}, \bibinfo{person}{Girish Sastry}, \bibinfo{person}{Heather Schmidt}, \bibinfo{person}{David Schnurr}, \bibinfo{person}{John Schulman}, \bibinfo{person}{Daniel Selsam},
  \bibinfo{person}{Kyla Sheppard}, \bibinfo{person}{Toki Sherbakov}, \bibinfo{person}{Jessica Shieh}, \bibinfo{person}{Sarah Shoker}, \bibinfo{person}{Pranav Shyam}, \bibinfo{person}{Szymon Sidor}, \bibinfo{person}{Eric Sigler}, \bibinfo{person}{Maddie Simens}, \bibinfo{person}{Jordan Sitkin}, \bibinfo{person}{Katarina Slama}, \bibinfo{person}{Ian Sohl}, \bibinfo{person}{Benjamin Sokolowsky}, \bibinfo{person}{Yang Song}, \bibinfo{person}{Natalie Staudacher}, \bibinfo{person}{Felipe~Petroski Such}, \bibinfo{person}{Natalie Summers}, \bibinfo{person}{Ilya Sutskever}, \bibinfo{person}{Jie Tang}, \bibinfo{person}{Nikolas Tezak}, \bibinfo{person}{Madeleine~B. Thompson}, \bibinfo{person}{Phil Tillet}, \bibinfo{person}{Amin Tootoonchian}, \bibinfo{person}{Elizabeth Tseng}, \bibinfo{person}{Preston Tuggle}, \bibinfo{person}{Nick Turley}, \bibinfo{person}{Jerry Tworek}, \bibinfo{person}{Juan Felipe~Cerón Uribe}, \bibinfo{person}{Andrea Vallone}, \bibinfo{person}{Arun Vijayvergiya}, \bibinfo{person}{Chelsea Voss},
  \bibinfo{person}{Carroll Wainwright}, \bibinfo{person}{Justin~Jay Wang}, \bibinfo{person}{Alvin Wang}, \bibinfo{person}{Ben Wang}, \bibinfo{person}{Jonathan Ward}, \bibinfo{person}{Jason Wei}, \bibinfo{person}{CJ Weinmann}, \bibinfo{person}{Akila Welihinda}, \bibinfo{person}{Peter Welinder}, \bibinfo{person}{Jiayi Weng}, \bibinfo{person}{Lilian Weng}, \bibinfo{person}{Matt Wiethoff}, \bibinfo{person}{Dave Willner}, \bibinfo{person}{Clemens Winter}, \bibinfo{person}{Samuel Wolrich}, \bibinfo{person}{Hannah Wong}, \bibinfo{person}{Lauren Workman}, \bibinfo{person}{Sherwin Wu}, \bibinfo{person}{Jeff Wu}, \bibinfo{person}{Michael Wu}, \bibinfo{person}{Kai Xiao}, \bibinfo{person}{Tao Xu}, \bibinfo{person}{Sarah Yoo}, \bibinfo{person}{Kevin Yu}, \bibinfo{person}{Qiming Yuan}, \bibinfo{person}{Wojciech Zaremba}, \bibinfo{person}{Rowan Zellers}, \bibinfo{person}{Chong Zhang}, \bibinfo{person}{Marvin Zhang}, \bibinfo{person}{Shengjia Zhao}, \bibinfo{person}{Tianhao Zheng}, \bibinfo{person}{Juntang Zhuang},
  \bibinfo{person}{William Zhuk}, {and} \bibinfo{person}{Barret Zoph}.} \bibinfo{year}{2024}\natexlab{}.
\newblock \bibinfo{title}{GPT-4 Technical Report}.
\newblock
\newblock
\showeprint[arxiv]{2303.08774}~[cs.CL]


\bibitem[Papineni et~al\mbox{.}(2002)]%
        {Papineni}
\bibfield{author}{\bibinfo{person}{Kishore Papineni}, \bibinfo{person}{Salim Roukos}, \bibinfo{person}{Todd Ward}, {and} \bibinfo{person}{Wei~Jing Zhu}.} \bibinfo{year}{2002}\natexlab{}.
\newblock \showarticletitle{BLEU: a Method for Automatic Evaluation of Machine Translation}.
\newblock  (\bibinfo{date}{10} \bibinfo{year}{2002}).
\newblock
\urldef\tempurl%
\url{https://doi.org/10.3115/1073083.1073135}
\showDOI{\tempurl}


\bibitem[Radford et~al\mbox{.}(2021)]%
        {radford2021learning}
\bibfield{author}{\bibinfo{person}{Alec Radford}, \bibinfo{person}{Jong~Wook Kim}, \bibinfo{person}{Chris Hallacy}, \bibinfo{person}{Aditya Ramesh}, \bibinfo{person}{Gabriel Goh}, \bibinfo{person}{Sandhini Agarwal}, \bibinfo{person}{Girish Sastry}, \bibinfo{person}{Amanda Askell}, \bibinfo{person}{Pamela Mishkin}, \bibinfo{person}{Jack Clark}, \bibinfo{person}{Gretchen Krueger}, {and} \bibinfo{person}{Ilya Sutskever}.} \bibinfo{year}{2021}\natexlab{}.
\newblock \bibinfo{title}{Learning Transferable Visual Models From Natural Language Supervision}.
\newblock
\newblock
\showeprint[arxiv]{2103.00020}~[cs.CV]


\bibitem[Reyes-Ortiz et~al\mbox{.}(2012)]%
        {misc_human_activity_recognition_using_smartphones_240}
\bibfield{author}{\bibinfo{person}{Jorge Reyes-Ortiz}, \bibinfo{person}{Davide Anguita}, \bibinfo{person}{Alessandro Ghio}, \bibinfo{person}{Luca Oneto}, {and} \bibinfo{person}{Xavier Parra}.} \bibinfo{year}{2012}\natexlab{}.
\newblock \bibinfo{title}{{Human Activity Recognition Using Smartphones}}.
\newblock \bibinfo{howpublished}{UCI Machine Learning Repository}.
\newblock
\newblock
\shownote{{DOI}: https://doi.org/10.24432/C54S4K}.


\bibitem[Su et~al\mbox{.}(2022)]%
        {10.1145/3517252}
\bibfield{author}{\bibinfo{person}{Jie Su}, \bibinfo{person}{Zhenyu Wen}, \bibinfo{person}{Tao Lin}, {and} \bibinfo{person}{Yu Guan}.} \bibinfo{year}{2022}\natexlab{}.
\newblock \showarticletitle{Learning Disentangled Behaviour Patterns for Wearable-based Human Activity Recognition}.
\newblock \bibinfo{journal}{\emph{Proc. ACM Interact. Mob. Wearable Ubiquitous Technol.}} \bibinfo{volume}{6}, \bibinfo{number}{1}, Article \bibinfo{articleno}{28} (\bibinfo{date}{mar} \bibinfo{year}{2022}), \bibinfo{numpages}{19}~pages.
\newblock
\urldef\tempurl%
\url{https://doi.org/10.1145/3517252}
\showDOI{\tempurl}


\bibitem[Tan and Bansal(2019)]%
        {tan2019lxmert}
\bibfield{author}{\bibinfo{person}{Hao Tan} {and} \bibinfo{person}{Mohit Bansal}.} \bibinfo{year}{2019}\natexlab{}.
\newblock \bibinfo{title}{LXMERT: Learning Cross-Modality Encoder Representations from Transformers}.
\newblock
\newblock
\showeprint[arxiv]{1908.07490}~[cs.CL]


\bibitem[Touvron et~al\mbox{.}(2023)]%
        {touvron2023llama}
\bibfield{author}{\bibinfo{person}{Hugo Touvron}, \bibinfo{person}{Thibaut Lavril}, \bibinfo{person}{Gautier Izacard}, \bibinfo{person}{Xavier Martinet}, \bibinfo{person}{Marie-Anne Lachaux}, \bibinfo{person}{Timothée Lacroix}, \bibinfo{person}{Baptiste Rozière}, \bibinfo{person}{Naman Goyal}, \bibinfo{person}{Eric Hambro}, \bibinfo{person}{Faisal Azhar}, \bibinfo{person}{Aurelien Rodriguez}, \bibinfo{person}{Armand Joulin}, \bibinfo{person}{Edouard Grave}, {and} \bibinfo{person}{Guillaume Lample}.} \bibinfo{year}{2023}\natexlab{}.
\newblock \bibinfo{title}{LLaMA: Open and Efficient Foundation Language Models}.
\newblock
\newblock
\showeprint[arxiv]{2302.13971}~[cs.CL]


\bibitem[Uddin and Soylu(2021)]%
        {uddin2021human}
\bibfield{author}{\bibinfo{person}{M.Z. Uddin} {and} \bibinfo{person}{A. Soylu}.} \bibinfo{year}{2021}\natexlab{}.
\newblock \showarticletitle{Human activity recognition using wearable sensors, discriminant analysis, and long short-term memory-based neural structured learning}.
\newblock \bibinfo{journal}{\emph{Sci Rep}}  \bibinfo{volume}{11} (\bibinfo{year}{2021}), \bibinfo{pages}{16455}.
\newblock
\urldef\tempurl%
\url{https://doi.org/10.1038/s41598-021-95947-y}
\showURL{%
\tempurl}


\bibitem[Vaswani et~al\mbox{.}(2023)]%
        {vaswani2023attention}
\bibfield{author}{\bibinfo{person}{Ashish Vaswani}, \bibinfo{person}{Noam Shazeer}, \bibinfo{person}{Niki Parmar}, \bibinfo{person}{Jakob Uszkoreit}, \bibinfo{person}{Llion Jones}, \bibinfo{person}{Aidan~N. Gomez}, \bibinfo{person}{Lukasz Kaiser}, {and} \bibinfo{person}{Illia Polosukhin}.} \bibinfo{year}{2023}\natexlab{}.
\newblock \bibinfo{title}{Attention Is All You Need}.
\newblock
\newblock
\showeprint[arxiv]{1706.03762}~[cs.CL]


\bibitem[Vedantam et~al\mbox{.}(2015)]%
        {vedantam2015cider}
\bibfield{author}{\bibinfo{person}{Ramakrishna Vedantam}, \bibinfo{person}{C.~Lawrence Zitnick}, {and} \bibinfo{person}{Devi Parikh}.} \bibinfo{year}{2015}\natexlab{}.
\newblock \bibinfo{title}{CIDEr: Consensus-based Image Description Evaluation}.
\newblock
\newblock
\showeprint[arxiv]{1411.5726}~[cs.CV]


\bibitem[Venugopalan et~al\mbox{.}(2015)]%
        {venugopalan2015sequence}
\bibfield{author}{\bibinfo{person}{Subhashini Venugopalan}, \bibinfo{person}{Marcus Rohrbach}, \bibinfo{person}{Jeff Donahue}, \bibinfo{person}{Raymond Mooney}, \bibinfo{person}{Trevor Darrell}, {and} \bibinfo{person}{Kate Saenko}.} \bibinfo{year}{2015}\natexlab{}.
\newblock \bibinfo{title}{Sequence to Sequence -- Video to Text}.
\newblock
\newblock
\showeprint[arxiv]{1505.00487}~[cs.CV]


\bibitem[Wang et~al\mbox{.}(2016)]%
        {wang2016hierarchical}
\bibfield{author}{\bibinfo{person}{Yilin Wang}, \bibinfo{person}{Suhang Wang}, \bibinfo{person}{Jiliang Tang}, \bibinfo{person}{Neil O'Hare}, \bibinfo{person}{Yi Chang}, {and} \bibinfo{person}{Baoxin Li}.} \bibinfo{year}{2016}\natexlab{}.
\newblock \bibinfo{title}{Hierarchical Attention Network for Action Recognition in Videos}.
\newblock
\newblock
\showeprint[arxiv]{1607.06416}~[cs.CV]


\bibitem[Wu et~al\mbox{.}(2020)]%
        {wu2020power}
\bibfield{author}{\bibinfo{person}{Kaishun Wu}, \bibinfo{person}{Yandao Huang}, \bibinfo{person}{Wenqiang Chen}, \bibinfo{person}{Lin Chen}, \bibinfo{person}{Xinyu Zhang}, \bibinfo{person}{Lu Wang}, {and} \bibinfo{person}{Rukhsana Ruby}.} \bibinfo{year}{2020}\natexlab{}.
\newblock \showarticletitle{Power saving and secure text input for commodity smart watches}.
\newblock \bibinfo{journal}{\emph{IEEE Transactions on Mobile Computing}} \bibinfo{volume}{20}, \bibinfo{number}{6} (\bibinfo{year}{2020}), \bibinfo{pages}{2281--2296}.
\newblock


\bibitem[Xsens(2022)]%
        {XsensMTwAwinda}
\bibfield{author}{\bibinfo{person}{Xsens}.} \bibinfo{year}{2022}\natexlab{}.
\newblock \bibinfo{title}{MTw Awinda wearable body-tracking system}.
\newblock \bibinfo{howpublished}{\url{https://www.xsens.com/products/mtw-awinda}}.
\newblock


\bibitem[Xu et~al\mbox{.}(2021)]%
        {xu2021e2evlp}
\bibfield{author}{\bibinfo{person}{Haiyang Xu}, \bibinfo{person}{Ming Yan}, \bibinfo{person}{Chenliang Li}, \bibinfo{person}{Bin Bi}, \bibinfo{person}{Songfang Huang}, \bibinfo{person}{Wenming Xiao}, {and} \bibinfo{person}{Fei Huang}.} \bibinfo{year}{2021}\natexlab{}.
\newblock \bibinfo{title}{E2E-VLP: End-to-End Vision-Language Pre-training Enhanced by Visual Learning}.
\newblock
\newblock
\showeprint[arxiv]{2106.01804}~[cs.CV]


\bibitem[Xu et~al\mbox{.}(2023)]%
        {xu2023mplug2}
\bibfield{author}{\bibinfo{person}{Haiyang Xu}, \bibinfo{person}{Qinghao Ye}, \bibinfo{person}{Ming Yan}, \bibinfo{person}{Yaya Shi}, \bibinfo{person}{Jiabo Ye}, \bibinfo{person}{Yuanhong Xu}, \bibinfo{person}{Chenliang Li}, \bibinfo{person}{Bin Bi}, \bibinfo{person}{Qi Qian}, \bibinfo{person}{Wei Wang}, \bibinfo{person}{Guohai Xu}, \bibinfo{person}{Ji Zhang}, \bibinfo{person}{Songfang Huang}, \bibinfo{person}{Fei Huang}, {and} \bibinfo{person}{Jingren Zhou}.} \bibinfo{year}{2023}\natexlab{}.
\newblock \bibinfo{title}{mPLUG-2: A Modularized Multi-modal Foundation Model Across Text, Image and Video}.
\newblock
\newblock
\showeprint[arxiv]{2302.00402}~[cs.CV]


\bibitem[Xu et~al\mbox{.}(2024)]%
        {10.1145/3643540}
\bibfield{author}{\bibinfo{person}{Xuhai Xu}, \bibinfo{person}{Bingsheng Yao}, \bibinfo{person}{Yuanzhe Dong}, \bibinfo{person}{Saadia Gabriel}, \bibinfo{person}{Hong Yu}, \bibinfo{person}{James Hendler}, \bibinfo{person}{Marzyeh Ghassemi}, \bibinfo{person}{Anind~K. Dey}, {and} \bibinfo{person}{Dakuo Wang}.} \bibinfo{year}{2024}\natexlab{}.
\newblock \showarticletitle{Mental-LLM: Leveraging Large Language Models for Mental Health Prediction via Online Text Data}.
\newblock \bibinfo{journal}{\emph{Proc. ACM Interact. Mob. Wearable Ubiquitous Technol.}} \bibinfo{volume}{8}, \bibinfo{number}{1}, Article \bibinfo{articleno}{31} (\bibinfo{date}{mar} \bibinfo{year}{2024}), \bibinfo{numpages}{32}~pages.
\newblock
\urldef\tempurl%
\url{https://doi.org/10.1145/3643540}
\showDOI{\tempurl}


\bibitem[Ye et~al\mbox{.}(2022)]%
        {ye2022hierarchical}
\bibfield{author}{\bibinfo{person}{Hanhua Ye}, \bibinfo{person}{Guorong Li}, \bibinfo{person}{Yuankai Qi}, \bibinfo{person}{Shuhui Wang}, \bibinfo{person}{Qingming Huang}, {and} \bibinfo{person}{Ming-Hsuan Yang}.} \bibinfo{year}{2022}\natexlab{}.
\newblock \bibinfo{title}{Hierarchical Modular Network for Video Captioning}.
\newblock
\newblock
\showeprint[arxiv]{2111.12476}~[cs.CV]


\bibitem[Yin et~al\mbox{.}(2023)]%
        {yin2023survey}
\bibfield{author}{\bibinfo{person}{Shukang Yin}, \bibinfo{person}{Chaoyou Fu}, \bibinfo{person}{Sirui Zhao}, \bibinfo{person}{Ke Li}, \bibinfo{person}{Xing Sun}, \bibinfo{person}{Tong Xu}, {and} \bibinfo{person}{Enhong Chen}.} \bibinfo{year}{2023}\natexlab{}.
\newblock \bibinfo{title}{A Survey on Multimodal Large Language Models}.
\newblock
\newblock
\showeprint[arxiv]{2306.13549}~[cs.CV]


\bibitem[Zhang et~al\mbox{.}(2023)]%
        {zhang2023videollama}
\bibfield{author}{\bibinfo{person}{Hang Zhang}, \bibinfo{person}{Xin Li}, {and} \bibinfo{person}{Lidong Bing}.} \bibinfo{year}{2023}\natexlab{}.
\newblock \bibinfo{title}{Video-LLaMA: An Instruction-tuned Audio-Visual Language Model for Video Understanding}.
\newblock
\newblock
\showeprint[arxiv]{2306.02858}~[cs.CL]


\bibitem[Zhao et~al\mbox{.}(2018)]%
        {10.1145/3230543.3230579}
\bibfield{author}{\bibinfo{person}{Mingmin Zhao}, \bibinfo{person}{Yonglong Tian}, \bibinfo{person}{Hang Zhao}, \bibinfo{person}{Mohammad~Abu Alsheikh}, \bibinfo{person}{Tianhong Li}, \bibinfo{person}{Rumen Hristov}, \bibinfo{person}{Zachary Kabelac}, \bibinfo{person}{Dina Katabi}, {and} \bibinfo{person}{Antonio Torralba}.} \bibinfo{year}{2018}\natexlab{}.
\newblock \showarticletitle{RF-based 3D skeletons}. In \bibinfo{booktitle}{\emph{Proceedings of the 2018 Conference of the ACM Special Interest Group on Data Communication}} (Budapest, Hungary) \emph{(\bibinfo{series}{SIGCOMM '18})}. \bibinfo{publisher}{Association for Computing Machinery}, \bibinfo{address}{New York, NY, USA}, \bibinfo{pages}{267–281}.
\newblock
\showISBNx{9781450355674}
\urldef\tempurl%
\url{https://doi.org/10.1145/3230543.3230579}
\showDOI{\tempurl}


\end{thebibliography}


\end{document}